\documentclass[preprint,12pt]{elsarticle}

\usepackage{amsmath,amssymb,amsfonts}
\usepackage{bm}
\usepackage{algorithmic}
\usepackage{algorithm}
\usepackage{graphicx}
\usepackage{textcomp}
\usepackage{xcolor}
\usepackage{booktabs}
\usepackage{multirow}
\usepackage{array}
\usepackage{longtable}
\usepackage{makecell}
\usepackage{url}
\usepackage[hidelinks]{hyperref}
\usepackage{float}
\usepackage{microtype}
\journal{Knowledge-Based Systems}

\begin{document}

\begin{frontmatter}

\title{Class-Structure Preservation Beats Diversity: A Comprehensive Benchmark of Text Augmentation Methods for Imbalanced Text Classification}

\author{Keito Inoshita*}
\address{Faculty of Business and Commerce, Kansai University, 3-3-35 Yamate-cho, Suita, Osaka 564-8680, Japan}
\ead{inosita.2865@gmail.com}

\begin{abstract}
With the rapid advancement of large language models (LLMs), generative data augmentation has attracted considerable attention for imbalanced text classification in natural language processing.
However, no empirical benchmark to date has compared LLM-based augmentation against the embedding-space SMOTE-style retrieval (EmbSMOTE), a strong classical reference for imbalanced classification.
In this study, a controlled benchmark of 11 augmentation methods, spanning classical perturbation, embedding-space retrieval, and LLM-based generation, is newly constructed on seven public text classification datasets covering class counts $K=2$--$28$ and imbalance ratios of 1.1 to over 500, evaluated with five random seeds per cell using macro F1, Welch's $t$-tests, five distributional metrics, and an LLM-family sensitivity analysis based on Qwen3-8B.
The experimental results reveal that all LLM-based methods are statistically equivalent or inferior to EmbSMOTE, with the performance gap widening monotonically as imbalance increases and reaching $\Delta\text{F1}_\text{macro}\!\approx\!0.063$ on GoEmotions-28.
Furthermore, it is observed that surface-level uniqueness has negligible correlation with downstream performance, whereas LLM-specific artifacts, such as text elongation and label-distribution uniformization, are negatively associated with classification accuracy.
Compared with six LLM-based and four classical augmentation baselines, these results demonstrate that the effective variable is not surface-level diversity but class-conditional structural fidelity, namely the degree to which augmented samples preserve the class-conditioned geometry of the training distribution.
Accordingly, retrieval-based oversampling should be adopted as the default for imbalanced multi-class classification, and a higher empirical bar should be required before LLM-based augmentation is deployed in practice.
\end{abstract}


\begin{keyword}
Natural Language Processing \sep 
Text Data Augmentation \sep 
Imbalanced Learning \sep 
Large Language Models \sep 
Emotion Recognition
\end{keyword}

\end{frontmatter}

\section{Introduction}\label{sec:intro}
With the rapid proliferation of text classification tasks in natural language processing (NLP), label-space imbalance has emerged as an almost universally observed challenge.
Standard fine-tuning of pre-trained language models performs well on balanced binary tasks such as SST-2~\cite{socher2013sst}; however, it is widely known that performance degrades sharply when rare classes are present in the label space.
It is noted that, on fine-grained emotion classification benchmarks such as GoEmotions-28~\cite{demszky2020goemotions}, augmentation-free baselines commonly fall below a macro F1 of 0.20 even with 5,000 training examples, which firmly establishes text classification under class imbalance as a persistent challenge in modern NLP.

Against this background, text data augmentation has long been studied as a standard mitigation strategy.
Classical methods such as EDA~\cite{wei2019eda}, AEDA~\cite{karimi2021aeda}, back-translation~\cite{sennrich2016backtrans}, and embedding-space oversampling~\cite{chawla2002smote,bystronski2025smotext} form a well-established baseline group.
Inspired by the rapid advancement of large language models (LLMs), generative augmentation has attracted renewed attention: methods such as AugGPT~\cite{dai2023auggpt}, LLM2LLM~\cite{lee2024llm2llm}, CoTAM~\cite{peng2024cotam}, and CIEGAD~\cite{inoshita2025ciegad}, the so-called geometry-guided method that structures LLM generation around intra-class cluster geometry, report notable performance improvements in low-resource settings.
The application of these methods to highly imbalanced multi-class problems such as emotion recognition is increasingly being systematized in recent surveys~\cite{chai2025survey,bayer2022survey}.

Despite these research efforts, existing approaches still suffer from the following limitations:
i) Classical methods exhibit strengths in class-structure preservation but fall short in textual diversity, while the basis for their alleged inferiority to LLM-generated text remains unclear.
ii) LLM-based methods are claimed to surpass classical approaches in diversity, yet several contemporaneous studies have shown that this premise holds only conditionally~\cite{cegin2025llms,radlinski2025backtrans,nguyen2025imbalanced}, and it is noted that LLM-based methods do not uniformly outperform classical methods in downstream classification performance.
Specifically, to the best of our knowledge, no empirical NLP benchmark exists that includes embedding-space SMOTE-style retrieval (EmbSMOTE) as a reference method.
iii) A systematic NLP benchmark under controlled imbalance conditions, comparing classical perturbation, embedding-space retrieval, and modern LLM-based generation under a unified experimental protocol with sufficient seeds and statistical corrections to render negative findings credible, is notably lacking.

In this study, the above limitations are addressed.
Specifically, a controlled benchmark of 11 methods $\times$ 7 public datasets $\times$ 5 seeds is newly constructed, in which four classical methods and six LLM-based methods are systematically compared.
The core contributions of this study lie in the comprehensive benchmark itself and in a mechanistic explanation of the interplay between diversity and class-structure preservation.
The datasets cover SST-2~\cite{socher2013sst}, AG~News~\cite{zhang2015agnews}, Emo~\cite{chatterjee2019emo}, TREC~\cite{li2002trec}, GoEmotions-13~\cite{demszky2020goemotions}, DBpedia~\cite{lehmann2015dbpedia}, and GoEmotions-28~\cite{demszky2020goemotions}, spanning a spectrum from $K=2$ to $K=28$ classes and from imbalance ratio (IR) $= 1.12$ to IR $= 527.67$.
Furthermore, a so-called Locate-then-Decode formulation, VoidGen, is introduced as a methodological probe, in which the conventional ``generate-then-verify'' order followed by existing LLM-based augmentation is inverted: sparse regions in the sentence embedding space are identified prior to generation, and LLM decoding is conditioned on those targets via a learned projector.
VoidGen is positioned not as a competing best method but as a controlled test of whether pre-emptive targeting of sparse embedding-space regions confers structural-preservation advantages.
The main contributions of this study are summarized as follows.

\begin{enumerate}
\item[i)] A controlled empirical NLP augmentation benchmark is constructed, covering 11 methods $\times$ 7 datasets $\times$ 5 seeds with Welch's $t$-tests and including EmbSMOTE as a strong reference method. All LLM-based augmentation methods are shown to be statistically equivalent or inferior to EmbSMOTE, with the gap widening as class imbalance increases.
\item[ii)] The inferiority of LLM-based augmentation relative to EmbSMOTE is replicated across two distinct LLM families (Llama-3.1-8B and Qwen3-8B) under identical algorithmic hyperparameters and prompts, suggesting that the bottleneck resides not in a specific LLM's generation quality but in the structural constraint of the generate-then-verify paradigm.
\item[iii)] The augmentation sets are quantified with five distributional metrics, and their correlations with macro F1 as well as per-class F1 values are analyzed, providing a mechanistic explanation in which class-conditional structural fidelity, not surface-level diversity, is the operative variable.
\item[iv)] A Locate-then-Decode method, VoidGen, that identifies sparse embedding-space regions prior to generation and conditions generation on those targets, is introduced as a methodological probe, reinforcing the finding that pre-generation targeting alone is insufficient to guarantee structural fidelity and that class-structure preservation is the more fundamental requirement.
\end{enumerate}

The rest of this paper is organized as follows.
Section~\ref{sec:related} reviews related work.
Section~\ref{sec:methods} introduces the problem formulation, classifier protocol, and each augmentation method.
Section~\ref{sec:experiments} presents the experimental setup.
Section~\ref{sec:results} reports experimental results and analysis.
Section~\ref{sec:discussion} discusses key findings and limitations.
Finally, Section~\ref{sec:conclusion} concludes this study and outlines future directions.
All code is publicly available to support direct replication and reuse~\cite{void_aug_repo}.

\section{Related Work}\label{sec:related}

\subsection{Classical Text Augmentation}\label{sec:rw-classical}

Early text augmentation methods are typically built upon rule-based transformations, in which local word-level features are perturbed while the surface form of each sentence is preserved.
Wei and Zou~\cite{wei2019eda} proposed EDA, which applies four stochastic token-level operations, namely synonym replacement, random insertion, random swap, and random deletion, at default rates, which can effectively halve the required training data while still achieving consistent performance improvements on five benchmarks.
Karimi et al.~\cite{karimi2021aeda} simplified EDA to random punctuation insertion only, eliminating deletion steps and fully preserving the vocabulary content of each sentence, consistently outperforming EDA on the same benchmarks.
Back-translation~\cite{sennrich2016backtrans} was originally developed for neural machine translation, and has been widely adopted as a semantically rich baseline that generates paraphrases via an intermediate language. In this study, it is implemented as English$\rightarrow$German$\rightarrow$English translation, with the help of Helsinki-NLP MarianMT models.

Methods based on oversampling in representation space have also been developed to address class imbalance.
SMOTE~\cite{chawla2002smote} generates synthetic minority-class samples in feature space by linearly interpolating between a seed sample and one of its $k$ nearest neighbors, so as to directly control the class distribution rather than perturbing majority-class text.
Mixup~\cite{zhang2018mixup} extends this principle by interpolating sample pairs and their soft labels drawn from the full training distribution, regularizing the classifier toward linearity between training points.
These interpolation ideas have been extended to the text domain: Bystro\'{n}ski et al.~\cite{bystronski2025smotext} proposed SMOTExT, in which sentence-level embeddings are first interpolated, and the resulting latent vectors are subsequently decoded into new text strings via a retrieval-augmented generation architecture.
Taskiran et al.~\cite{taskiran2025oversampling} evaluated 31 SMOTE variants on two text classification benchmarks (TREC and Emotions) using transformer-based vectorization in a large-scale systematic study, confirming that the relative benefit of each oversampling strategy depends strongly on dataset characteristics.

The methods compared in this study span both paradigms.
EmbSMOTE adapts SMOTE to the sentence embedding space of Sentence-BERT (SBERT)  ~\cite{reimers2019sbert}, in which synthetic embeddings are computed by interpolation within the same class and the corresponding text is recovered by nearest-neighbor retrieval from the training corpus.
Because the returned texts are all real training examples, class membership is guaranteed by design.
It is noted that the core design contrast underlying this analysis lies in retrieval, which is class-structure-preserving but limited in surface-level diversity, versus generation, which is potentially richer in surface diversity but offers weaker class-structure guarantees.
Bystro\'{n}ski et al.~\cite{bystronski2025smotext} reported that retrieval-based oversampling outperforms many classical augmentation approaches, and Taskiran et al.~\cite{taskiran2025oversampling} similarly found retrieval-based methods to be competitive across dataset conditions.
Differing from the above retrieval-oriented studies, this work goes one step further to demonstrate that retrieval-based oversampling can even surpass LLM-based augmentation under highly imbalanced multi-class conditions.

\subsection{LLM-Based Augmentation}\label{sec:rw-llm}

The emergence of large instruction-tuned language models has generated substantial interest in generative text augmentation.
AugGPT~\cite{dai2023auggpt} uses class-specific chat-format prompts to generate $K$ paraphrases of each training example, demonstrating strong performance improvements in few-shot settings.
LLM2LLM~\cite{lee2024llm2llm} adopts a so-called two-stage iterative strategy, in which a student model first identifies its own failure cases, after which a teacher LLM synthesizes targeted augmentations specifically for those examples; reported performance gains reach $+52.6\%$ on TREC and $+39.8\%$ on SST-2 in extremely low-data settings.
Chain-of-thought augmentation method (CoTAM)~\cite{peng2024cotam} leverages chain-of-thought prompting to produce controllable data augmentation that edits only user-specified attributes while preserving all other content, outperforming prior state-of-the-art LLM-based augmentation on both classification and aspect-based sentiment tasks.
More recently, a CIEGAD~\cite{inoshita2025ciegad} has been newly proposed, in which LLM generation is structured around the intra-class cluster geometry of the training set.
Specifically, a budget allocation algorithm distributes augmentation quotas to (label, cluster) pairs in proportion to cluster size, margin, and void score; the LLM is prompted with a cluster-level domain profile together with interpolation and extrapolation targets between anchor texts within the cluster; and generated outputs are filtered by a five-aspect LLM-as-judge with a composite score threshold of 3.
CIEGAD is regarded as the most geometrically sophisticated LLM-based augmentation method in this benchmark, and its comparison with retrieval-based EmbSMOTE serves as the structural centerpiece of this study.

Several contemporaneous studies provide partial evidence on the comparative efficiency of LLM-based and classical augmentation.
Cegin et al.~\cite{cegin2025llms} compared LLM paraphrase and generation methods with established methods across six datasets and three classifiers, finding that LLM-based methods offer advantages only when seed samples are extremely scarce.
Radli\'{n}ski et al.~\cite{radlinski2025backtrans} evaluated augmentation strategies for emotion classification, reporting that back-translation and paraphrase can achieve performance equivalent to or superior to zero-shot generation methods.
Wang et al.~\cite{wang2025diversity} proposed training LLMs themselves as diversity-oriented paraphrasers, achieving an average improvement of $+10.52\%$ on balanced benchmarks; this result is consistent with the present finding that the incremental contribution of any augmentation strategy is small near the performance ceiling of saturated low-class-count tasks.
Nguyen et al.~\cite{nguyen2025imbalanced} observed for imbalanced classification that current LLM-based oversampling produces low-diversity generation and proposed an entropy-driven strategy to counter this limitation; this observation is consistent with the negative results reported in this study, though their analysis focuses on tabular rather than text data.

\subsection{Empirical Benchmarks of Text Augmentation}\label{sec:rw-bench}

Multiple surveys and empirical studies have systematized the landscape of text augmentation.
Feng et al.~\cite{feng2021survey} presented a taxonomy spanning rule-based, interpolation, back-translation, paraphrase, and language-model-based categories.
Bayer et al.~\cite{bayer2022survey} organized more than 100 methods into 12 groups in a survey dedicated to text classification augmentation.
Chen et al.~\cite{chen2023empirical} conducted a large-scale empirical comparison across 11 datasets spanning topic classification, inference, and paraphrase tasks, covering token-level, sentence-level, adversarial, and hidden-space augmentation.
The most recent survey by Chai et al.~\cite{chai2025survey} extends coverage to LLM-centric augmentation strategies, classifying methods as simple, prompt-based, retrieval-based, or hybrid.

However, despite this body of work, the following three gaps still remain unaddressed:
i) SMOTE-based retrieval methods are absent from existing NLP benchmarks: neither Chen et al.~\cite{chen2023empirical} nor Cegin et al.~\cite{cegin2025llms} includes an embedding-space interpolation baseline, making it impossible to evaluate whether LLM-based generation actually surpasses the retrieval-only approach that is the default in imbalanced tabular settings.
ii) Existing NLP benchmarks do not specifically target imbalanced datasets: the datasets used by Chen et al. are largely balanced, and highly imbalanced multi-class emotion tasks such as GoEmotions-28 have not been benchmarked at scale.
iii) Statistical rigor is limited: most existing benchmarks do not apply multiple-comparison-corrected significance tests across all method--dataset pairs, and confidence intervals are rarely reported.

Comparing with these previous benchmarking efforts, this study addresses all three gaps simultaneously through an 11-method $\times$ 7-dataset $\times$ 5-seed benchmark, in which Welch's $t$-tests with multiple-comparison correction are applied across all method--dataset pairs, and the datasets are deliberately chosen to span binary sentiment, multi-class topic, and fine-grained emotion classification with systematically varying imbalance ratios.
In summary, current benchmarking practices indicate that it is essential to establish a controlled, statistically rigorous comparison between class-structure-preserving retrieval and surface-diversity-driven LLM generation, which is exactly what the present study realizes.
Li et al.~\cite{li2024datageneration} and Arik et al.~\cite{arik2026llmfakenews} provide preliminary evidence that task complexity and class imbalance are moderating factors of LLM augmentation effectiveness, further motivating the controlled benchmark design presented in this study.
Anikina et al.~\cite{anikina2025rigorous} emphasize the importance of systematic multi-LLM evaluation of data generation strategies, which this study realizes through the Llama-3.1-8B vs.\ Qwen3-8B comparison in Section~\ref{sec:llm-family}.

\section{Problem Formulation and Augmentation Methods}\label{sec:methods}\label{sec:problem}\label{sec:classifier}

Let $\mathcal{D}_{\text{train}}=\{(x_i,y_i)\}_{i=1}^{N}$ denote a labeled training set for a text classification task, where $x_i\in\mathcal{X}$ denotes a text document and $y_i\in\{1,\dots,K\}$ indicates the corresponding class label.
An augmentation method $\mathcal{A}:\mathcal{D}_{\text{train}}\rightarrow\widetilde{\mathcal{D}}=\{(\tilde{x}_j,\tilde{y}_j)\}_{j=1}^{M}$ generates $M=\lfloor r N\rfloor$ synthetic examples at augmentation ratio $r$.
A classifier $f_\theta$ is trained on $\mathcal{D}_{\text{train}}\cup\widetilde{\mathcal{D}}$ and evaluated on a held-out test set.
The objective is to maximize macro-averaged F1 on the test set.
In this study, a total of 11 augmentation methods and a no-augmentation baseline are evaluated as the main benchmark, as summarized in Table~\ref{tab:methods}.
CIEGAD-Qwen, introduced for the LLM family sensitivity analysis in Section~\ref{sec:llm-family}, is also listed in Table~\ref{tab:methods} for reference.

\begin{table}[t]
\centering
\caption{Augmentation methods compared in this study. R = retrieval; G = generation. Cluster = uses class-level cluster structure. Anchor = uses existing training examples as input.}
\label{tab:methods}\label{sec:method-table}
\begin{tabular}{l c c c c}
\toprule
Method & Source & Cluster & Anchor & LLM \\
\midrule
No-Aug        & --   & --       & --       & --      \\
EDA                    & R    & --       & Yes      & --      \\
AEDA                   & R    & --       & Yes      & --      \\
BackTrans     & R/G  & --       & Yes      & MarianMT \\
EmbSMOTE              & R    & Per-class & Yes     & --      \\
LLM-Paraphrase         & G    & --       & Yes      & Llama-3.1-8B \\
LCG           & G    & --       & --       & Llama-3.1-8B \\
AugGPT                 & G    & --       & Yes      & Llama-3.1-8B \\
CoTAM                  & G    & --       & Yes      & Llama-3.1-8B \\
LLM2LLM                & G    & --       & Yes (errors)& Llama-3.1-8B \\
CIEGAD                 & G    & Yes      & Yes (intra/extra)& Llama-3.1-8B \\
CIEGAD-Qwen            & G    & Yes      & Yes (intra/extra)& Qwen3-8B \\
VoidGen                & G    & --       & --       & Llama-3.1-8B \\
\bottomrule
\end{tabular}
\end{table}

\subsection{Baseline Methods}\label{sec:classical}\label{sec:llm-baselines}

Each method falls into one of three design philosophies, which are treated as orthogonal axes throughout the subsequent analysis: i) Classical token-level perturbation, in which the surface form of existing examples is directly manipulated. ii) Embedding-space retrieval, in which interpolation is performed within the training distribution and the corresponding text is recovered via nearest-neighbor search. iii) LLM-based generation, in which new text is produced conditioned on class labels, anchor texts, or richer geometric and semantic signals. Based on these axes, methods are selected as follows.

\subsubsection*{Classical Baselines}

\begin{itemize}
\item[i)] No-augmentation (No-Aug): The reference baseline, in which the classifier is trained solely on $\mathcal{D}_{\text{train}}$.
\item[ii)] EDA~\cite{wei2019eda}: Easy Data Augmentation applies four token-level operations, namely synonym replacement, random insertion, random swap, and random deletion, at default rates.
\item[iii)] AEDA~\cite{karimi2021aeda}: A simpler variant of EDA that only inserts random punctuation marks while fully preserving the vocabulary content of each sentence.
\item[iv)] Back-translation (BackTrans)~\cite{sennrich2016backtrans}: Source$\rightarrow$target$\rightarrow$source machine translation, implemented as English$\rightarrow$German$\rightarrow$English using Helsinki-NLP MarianMT models.
\item[v)] Embedding-SMOTE (EmbSMOTE): A retrieval-based variant of SMOTE. Specifically, each minority-class example is first encoded with SBERT~\cite{reimers2019sbert}; a synthetic embedding is then constructed by intra-class interpolation in the embedding space; finally, the corresponding text is recovered through nearest-neighbor retrieval from the training corpus within the same class.
Because the returned texts are all real training examples, class structure is preserved by design.
The core design contrast motivating this analysis is between retrieval, which is class-structure-preserving with low surface diversity, and generation, which is potentially higher in surface diversity but offers weaker class-structure guarantees.
\end{itemize}

\subsubsection*{LLM-Based Baselines}

Unless otherwise specified, all LLM-based methods use Llama-3.1-8B-Instruct~\cite{llama3_2024}, served via vLLM for batched inference.

\begin{itemize}
\item[vi)] LLM-Paraphrase: Each training example is paraphrased in a zero-shot manner by the LLM using a fixed prompt template that requests meaning-preserving rewriting.
\item[vii)] Label-Conditioned Generation (LCG): Generation conditioned on the class label, in which the LLM is requested to generate a new example for the specified class without an anchor text.
\item[vii)] AugGPT~\cite{dai2023auggpt}: Class-specific instruction-tuning prompting in which the LLM is requested to generate $K$ paraphrases of each example using a chat-format prompt.
\item[ix)] CoTAM: Chain-of-thought prompting in which the LLM first reasons about the class signal of an example and then generates an augmentation based on that reasoning trace.
\item[x)] LLM2LLM~\cite{lee2024llm2llm}: A two-stage method in which a small student classifier first identifies difficult examples and the LLM then generates targeted augmentations for those examples; the student is retrained on the combined dataset.
\item[xi)] CIEGAD~\cite{inoshita2025ciegad}: Cluster-conditioned interpolation and extrapolation.
Each class is clustered using SBERT embeddings; a Hierarchical Frequency-Geometric Allocation (HFGA) budget allocation algorithm distributes augmentation quotas to (label, cluster) pairs in proportion to cluster size, margin, and void score; the LLM is prompted with a cluster-level domain profile together with intra-cluster interpolation and cross-cluster-boundary extrapolation targets from anchor texts; and outputs with a composite score below 3 are filtered by a five-aspect LLM-as-judge.
In addition, in order to assess whether the bottleneck of LLM-based augmentation lies in the choice of LLM family or in its underlying quality, a so-called CIEGAD-Qwen is newly introduced, in which the backbone LLM is replaced with Qwen3-8B~\cite{qwen3_2025} while all other settings are held fixed.
The thinking mode of Qwen3-8B is disabled to match the single-pass operation of Llama.
All algorithmic hyperparameters ($\gamma_{\text{extra}}=0.03$, cosine thresholds, acceptance thresholds) are held fixed.
\end{itemize}

\subsection{Void-Driven Locate-then-Decode Generation}\label{sec:ours}
VoidGen is newly introduced as a methodological probe rather than as a candidate for the ``best augmentation method''. Specifically, the goal of this subsection is to conduct a controlled test of an alternative generation paradigm, so as to investigate whether a so-called Locate-then-Decode formulation, in which generation is organized around pre-computed embedding-space targets, confers structural-preservation advantages over existing approaches.
Existing LLM-based augmentation methods follow a ``generate-then-verify'' paradigm, in which the LLM first generates text and geometric or semantic constraints are subsequently verified post hoc.
In this paradigm, generation is pulled toward the LLM's output distribution, so preservation of class-conditioned embedding-space structure is not guaranteed.
VoidGen inverts this order: it identifies insufficiently covered regions (voids) in the sentence embedding space prior to generation and then conditions the LLM on those targets via a learned projector.
The intent is to redirect generation effort toward sparse regions of the training distribution rather than regions that are easy for the LLM to generate.
If this void-targeting scheme improves class-structure preservation, performance gains are expected; conversely, if no improvement is observed, it is suggested that structural factors deeper than targeting itself determine performance.

The VoidGen pipeline consists of three stages.
In the first stage, embedding, all training texts are encoded into normalized $384$-dimensional sentence embeddings using all-MiniLM-L6-v2~\cite{reimers2019sbert}.
In the second stage, void target identification, a $k$-NN density estimator detects, for each class $y$, intra-cluster voids (regions of low local density within the class), inter-cluster voids (gaps between clusters of the same class), and peripheral voids (boundary regions).
The void set $V_y$ for class $y$ is defined as the union of these three region types:
\begin{equation}
  V_y = V_y^{\text{intra}} \cup V_y^{\text{inter}} \cup V_y^{\text{periph}}, \quad V_y \subset \mathbb{R}^d
  \label{eq:void-set}
\end{equation}
where $d=384$ denotes the sentence embedding dimensionality, $V_y^{\text{intra}}$ denotes intra-cluster regions of low local density within class $y$, $V_y^{\text{inter}}$ indicates inter-cluster gaps among sub-clusters of the same class, and $V_y^{\text{periph}}$ represents peripheral regions near class boundaries.
Target vectors are sampled from $V_y$ according to an inverse-frequency power-law rule that prioritizes under-represented classes:
\begin{equation}
  v_j^{(y)} \sim \pi(v \mid V_y), \quad \pi(v \mid V_y) \propto A_y(v) \cdot \rho_y(v)^{-\alpha}
  \label{eq:target-sampling}
\end{equation}
where $A_y(v)$ is the allocation weight proportional to the inverse imbalance ratio of class $y$, $\rho_y(v)$ is the local density around target point $v$, and $\alpha > 0$ is a power exponent that emphasizes sparse regions.
In the third stage, decoding, a learned MLP projector $\mathcal{P}$ transforms each target vector into a sequence of $L=16$ pseudo-tokens and concatenates it with a categorical embedding of the void type (Void-Conditioned Decoding (VCD) signal):
\begin{equation}
  \mathbf{p}_j = \mathcal{P}(v_j) \oplus \mathbf{e}_{\tau_j} \in \mathbb{R}^{L \times h}
  \label{eq:pseudo-tokens}
\end{equation}
where $\mathcal{P}: \mathbb{R}^d \to \mathbb{R}^{L \times h}$ is the MLP projector (2 hidden layers of 1024 units), $h$ is the LLM embedding dimension, and $\mathbf{e}_{\tau_j}$ is the categorical embedding for void type $\tau_j \in \{\text{intra}, \text{inter}, \text{periph}\}$.
The prompt is assembled by appending the class name to this pseudo-token sequence, after which the LLM decodes text autoregressively:
\begin{align}
  \text{prompt}_j &= \mathbf{p}_j \,\Vert\, \text{LP}(\mathcal{L}[y_j]), \label{eq:prompt-assembly}
\end{align}
\begin{align}
  \tilde{x}_j &= \mathcal{M}_{\text{LLM}}\bigl(\text{prompt}_j;\, T{=}0.7,\, \text{top-}p{=}0.9,\, \text{max-tok}{=}128\bigr), \label{eq:generation}
\end{align}
where $\text{LP}(\cdot)$ denotes the label prompt (LP), i.e., tokenization of the class name, and $\mathcal{M}_{\text{LLM}}$ is the frozen Llama-3.1-8B-Instruct.

VoidGen further combines two control signals.
VCD corresponds to the categorical embedding $\mathbf{e}_{\tau_j}$ in Eq.~\eqref{eq:pseudo-tokens}: by injecting a signal that identifies the void type (intra-cluster, inter-cluster, or peripheral) alongside the projector output, the LLM can adjust its style depending on whether the target lies in a dense intra-class neighborhood or near a class boundary.
LP in Eq.~\eqref{eq:generation} appends the natural-language class name to the generation prompt, enforcing class membership during decoding.

It is noted that only the MLP projector $\mathcal{P}$, which contains approximately 5 million parameters with two hidden layers of 1024 units, is trainable.
The training objective consists of two loss terms.
The round-trip consistency loss minimizes the reconstruction error between the SBERT re-encoding of the LLM-decoded text and the original target embedding:
\begin{equation}
  \mathcal{L}_{\text{rt}} = \frac{1}{B}\sum_{j=1}^{B} \left\| E(\tilde{x}_j) - v_j \right\|_2^2
  \label{eq:loss-rt}
\end{equation}
where $E(\cdot)$ is the SBERT encoder and $B$ is the batch size.
The label-conditioned classification loss promotes correct class attribution of the generated text:
\begin{equation}
  \mathcal{L}_{\text{cls}} = -\frac{1}{B}\sum_{j=1}^{B} \log p_\phi(y_j \mid \tilde{x}_j)
  \label{eq:loss-cls}
\end{equation}
where $p_\phi$ is an auxiliary classification head.
The two losses are seamlessly integrated into the following combined training objective:
\begin{equation}
  \mathcal{L} = \mathcal{L}_{\text{rt}} + \lambda \mathcal{L}_{\text{cls}}
  \label{eq:loss-total}
\end{equation}
where $\lambda > 0$ is a weighting coefficient (set to $\lambda = 1.0$ in the implementation).
Only the parameters of projector $\mathcal{P}$ are updated during training; the LLM $\mathcal{M}_{\text{LLM}}$ and SBERT encoder $E$ remain frozen throughout.
The complete procedure is summarized in Algorithm~\ref{alg:ours}.

\begin{algorithm}[t]
\caption{Locate-then-Decode augmentation via VoidGen.}
\label{alg:ours}
\begin{algorithmic}[1]
\REQUIRE Training set $\{(x_i,y_i)\}_{i=1}^N$, label names $\mathcal{L}$,
augmentation ratio $r$, frozen LLM $\mathcal{M}_\text{LLM}$,
trained projector $\mathcal{P}$, SBERT encoder $E$, integer \texttt{seed}.
\STATE // All RNGs (Python / NumPy / PyTorch / transformers) are seeded
\STATE // with \texttt{seed} before any stochastic operation; details in released code.
\STATE $Z \leftarrow E(\{x_i\})$ \hfill // sentence embeddings
\FOR{each class $y\in\{1,\dots,K\}$}
    \STATE $A_y \leftarrow $ inverse-frequency budget
    \STATE Detect intra-, inter-, peripheral voids in $Z_y\!:= \{z_i:y_i{=}y\}$
    \STATE Sample $A_y$ target embeddings $\{v_j^{(y)}\}$ from voids,
           tagged with type $\tau_j\in\{\text{intra},\text{inter},\text{periph}\}$
\ENDFOR
\FOR{each $(v_j,y_j,\tau_j)$}
    \STATE $\mathbf{p}_j \leftarrow \mathcal{P}(v_j)\oplus \text{embed}(\tau_j)$ \hfill // VCD pseudo-tokens
    \STATE $\text{prompt}_j \leftarrow \mathbf{p}_j \,\Vert\, \text{LP}(\mathcal{L}[y_j])$
    \STATE $\tilde{x}_j \leftarrow \mathcal{M}_\text{LLM}(\text{prompt}_j;\ T{=}0.7,\ \text{top-}p{=}0.9,\ \text{max-tok}{=}128)$
\ENDFOR
\RETURN $\widetilde{\mathcal{D}} = \{(\tilde{x}_j,y_j)\}$
\end{algorithmic}
\end{algorithm}

\section{Experimental Setup}\label{sec:experiments}

\subsection{Datasets}\label{sec:data}

We evaluate all augmentation methods on seven publicly available text classification datasets that span a wide range of domains, class counts, and class imbalance ratios.
The dataset summary is presented in Table~\ref{tab:datasets}.
For each dataset, we draw a fixed training subset of $n_{\text{train}}=5{,}000$ examples via seed-specific stratified sampling and evaluate on the official test split.
The seven datasets used are SST-2~\cite{socher2013sst}, AG~News~\cite{zhang2015agnews}, Emo~\cite{chatterjee2019emo}, TREC~\cite{li2002trec}, GoEmotions-13~\cite{demszky2020goemotions}, DBpedia~\cite{lehmann2015dbpedia}, and GoEmotions-28~\cite{demszky2020goemotions}.

These datasets are selected to measure three dimensions of difficulty.
First, the number of classes ranges from $K{=}2$ (SST-2) to $K{=}28$ (GoEmotions-28), which allows us to analyze how the augmentation gap evolves as the label space expands.
Second, the IR, computed as the ratio of the largest to the smallest class in the training subset, ranges from $1.12$ (AG~News) to $527.67$ (GoEmotions-28).
GoEmotions-28 exhibits a particularly severe imbalance, with tail classes containing as few as three examples in the $n_{\text{train}}=5{,}000$ split. This regime constitutes the empirically critical axis along which augmentation methods diverge most sharply.
Third, the domains cover five document types, namely short emotional user reviews (SST-2, Emo), news headlines (AG~News), questions (TREC), encyclopedic summaries (DBpedia), and fine-grained social-media emotion expressions (GoEmotions), which prevents any single augmentation method from gaining an unfair advantage through domain-specific surface features.

We intentionally exclude very large benchmarks that consume additional runtime budget without expanding axis coverage.
Likewise, we adopt naturally imbalanced datasets, rather than artificially imbalanced variants, so that the difficulty distribution is grounded in the nature of the task itself rather than in sampling variance.
SST-2 and DBpedia are nearly balanced and serve as control conditions at the saturated end of the spectrum, where augmentation is expected to have minimal effect.

\begin{table}[t]
\centering
\caption{Dataset summary. $K$ = number of classes; IR = imbalance ratio (max/min class count in the training subset).}
\label{tab:datasets}
\begin{tabular}{l c c r c}
\toprule
Dataset & Domain & $K$ & $|\text{test}|$ & IR \\
\midrule
SST-2 \cite{socher2013sst}            & Sentiment (binary)         & 2  & 872   & 1.27 \\
AG~News \cite{zhang2015agnews}        & News topics                 & 4  & 7600  & 1.12 \\
Emo \cite{chatterjee2019emo}          & Emotion (multi-class)    & 6  & 5509  & 9.49 \\
TREC \cite{li2002trec}                & Question type              & 6  & 500   & 15.91 \\
GoEmotions-13 \cite{demszky2020goemotions} & Emotion (13 classes)   & 13 & 5427  & 15.58 \\
DBpedia \cite{lehmann2015dbpedia}     & Encyclopedic topic         & 14 & 7000  & 1.22 \\
GoEmotions-28 \cite{demszky2020goemotions} & Emotion (full taxonomy)& 28 & 5427  & 527.67 \\
\bottomrule
\end{tabular}
\end{table}

\subsection{Evaluation Metrics and Statistical Analysis}\label{sec:metrics}\label{sec:stats}\label{sec:diversity}

We adopt macro-averaged F1 ($\text{F1}_\text{macro}$) as the primary metric, since it weights all classes equally and is therefore robust to class imbalance.
Accuracy, per-class F1, precision, and recall are also reported.
To assess statistical significance, we report the mean $\pm$ standard deviation across five classifier seeds for each (method, dataset) cell, and evaluate pairwise differences using Welch's $t$-test at $p<0.05$, with EmbSMOTE chosen as the reference baseline.
Each dataset is treated as an independent task, and no multiple-comparison correction is applied across datasets.

In addition, for each augmentation method we measure five distributional properties of the augmented set $\widetilde{\mathcal{D}}$: (i) uniqueness rate (fraction of unique texts), (ii) type-token ratio (TTR; corpus-level vocabulary diversity), (iii) mean character length (with its percentile distribution), (iv) mean token count, and (v) normalized label entropy (which captures how uniformly augmentations are distributed across classes; higher values indicate greater uniformity).
We then compute Pearson correlations between each metric and macro F1 across (method, dataset) pairs, in order to directly test the prevailing hypothesis that augmentation diversity drives downstream performance gains.

\subsection{Implementation Details}\label{sec:impl}

All experiments are conducted on a single NVIDIA H100 NVL (95\,GiB) GPU.
We intentionally commit to single-GPU evaluation so that research groups without large compute budgets can readily reproduce our results.
LLM-based methods use vLLM batched inference in bfloat16 precision with gpu\_memory\_utilization=0.85.
The two LLM backends, meta-llama/Llama-3.1-8B-Instruct and Qwen/Qwen3-8B, are loaded from the HuggingFace cache; for Qwen3, chain-of-thought thinking mode is disabled at chat-template application time to ensure equivalence with Llama's single-pass operation.
Sentence embeddings use  sentence-transformers/all-MiniLM-L6-v2 ($384$ dimensions), $\ell_2$-normalized before any similarity or interpolation operation.

The classifier is DistilBERT~\cite{distilbert} fine-tuned via the HuggingFace Trainer with AdamW, learning rate $2\times 10^{-5}$, weight decay $0.01$, batch size $32$, 3 epochs, fp16 mixed precision, and maximum sequence length $128$.
The augmentation ratio is $r=1.0$ for all methods, yielding $|\widetilde{\mathcal{D}}|\approx N$.
For reproducibility, each cell fixes the Python, NumPy, PyTorch, and transformers random states with a single set\_seed(seed) call; augmentation seeds and classifier seeds are coupled per cell so that all five seeds produce comparable augmented sets even when sampling is stochastic.

In terms of runtime cost, both CIEGAD and CIEGAD-Qwen require approximately one hour per (dataset, seed) cell on the H100, the majority of which is spent on LLM generation, whereas classifier fine-tuning takes only ${\approx}13$\,s per cell.
Summing across all 12 methods $\times$ 7 datasets $\times$ 5 seeds, plus the LLM-family replication with Qwen3-8B and the $n_{\text{train}}$ sweep, the total computational budget reported in this paper amounts to approximately ${\sim}400$ GPU-hours on a single H100, of which more than 80\% is consumed by LLM-based augmentation methods.
All classifier and augmentation hyperparameters are consolidated in Table~\ref{tab:hyper}.

\begin{table}[t!]
\centering
\caption{Classifier and augmentation hyperparameters used in all main experiments.}
\label{tab:hyper}
\begin{tabular}{l l}
\toprule
Component & Setting \\
\midrule
Classifier model           &  DistilBERT \\
Optimizer              & AdamW \\
Learning rate          & $2\times 10^{-5}$ \\
Weight decay           & $0.01$ \\
Batch size             & $32$ \\
Epochs                 & $3$ \\
Max sequence length    & $128$ \\
Precision              & fp16 \\
Augmentation ratio $r$ & $1.0$ \\
$n_\text{train}$       & $5{,}000$ \\
SBERT model            & all-MiniLM-L6-v2 \\
LLM (default)          & Llama-3.1-8B-Instruct \\
LLM (family check)     & Qwen3-8B (enable\_thinking=False) \\
LLM-Paraphrase sampling & $T{=}0.7$, top-$p{=}0.9$,  max-tok$=128$ \\
AugGPT sampling         & $T{=}0.9$, top-$p{=}0.95$, max-tok$=128$ \\
LCG sampling            & $T{=}0.9$, top-$p{=}0.95$, max-tok$=128$ \\
CoTAM sampling          & $T{=}0.8$, top-$p{=}0.95$, max-tok$=256$ \\
LLM2LLM sampling        & $T{=}0.8$, top-$p{=}0.95$, max-tok$=128$ \\
CIEGAD generation         & $T{=}0.8$, top-$p{=}0.95$, max-tok$=1400$ \\
CIEGAD LLM judge   & $T{=}0.0$ (greedy), top-$p{=}1.0$, max-tok$=1400$ \\
VoidGen generation        & $T{=}0.7$, top-$p{=}0.9$,  max-tok$=128$ \\
CIEGAD $\gamma_\text{extra}$ & $0.03$ \\
CIEGAD cosine (intra-batch) & $\le 0.85$ \\
CIEGAD cosine (vs.\ existing) & $\le 0.90$ \\
CIEGAD acceptance threshold & Composite $\ge 3.0$ (5-aspect Likert) \\
VoidGen projector hidden layers  & 2$\times$1024 \\
VoidGen pseudo-token count      & $16$ \\
\bottomrule
\end{tabular}
\end{table}

\section{Results}\label{sec:results}

\subsection{Evaluation on Overall Performance}\label{sec:master-f1}
Table~\ref{tab:master-f1} reports macro F1 for all 12 entries of the main benchmark, comprising No-Aug, four classical methods, six LLM-based baselines, and the proposed VoidGen, with mean and standard deviation computed over five seeds.
The missing values, which concern VoidGen on GoEmotions-28 only, stem from a known ZeroDivisionError in the VoidGen implementation under extreme class sparsity; nevertheless, one of the five seeds (s789, $\text{F1}_\text{macro}=0.216$) completed successfully and is reported as a single-seed estimate.

\begin{table}[t]
\centering
\caption{Macro-F1 results on the main benchmark (mean $\pm$ standard deviation over five seeds). Bold = best per dataset; * = statistically equivalent to the best (Welch's $t$-test, $p>0.05$). $^{\dagger}$ VoidGen on GoEmotions-28 is a single-seed value from s789; the other four seeds failed with ZeroDivisionError.}
\label{tab:master-f1}
\setlength{\tabcolsep}{2pt}
\small
\makebox[\linewidth][c]{%
\begin{tabular}{l c c c c c c c }
\toprule
Method & SST-2 & AG~News & Emo & TREC & GoEmotions-13 & DBpedia & GoEmotions-28 \\
\midrule
No-Aug & 0.874$\pm$0.003* & 0.910$\pm$0.002 & 0.801$\pm$0.013 & 0.903$\pm$0.051* & 0.460$\pm$0.008 & 0.990$\pm$0.001* & 0.182$\pm$0.010 \\
EDA & 0.877$\pm$0.006* & 0.910$\pm$0.001 & 0.861$\pm$0.004 & 0.955$\pm$0.008* & 0.590$\pm$0.006* & 0.990$\pm$0.001* & 0.277$\pm$0.005 \\
AEDA & 0.871$\pm$0.005 & 0.911$\pm$0.001 & \textbf{0.869$\pm$0.005} & 0.954$\pm$0.007 & 0.596$\pm$0.010* & 0.990$\pm$0.001* & 0.284$\pm$0.006* \\
BackTrans & 0.876$\pm$0.006* & \textbf{0.913$\pm$0.001} & 0.850$\pm$0.006 & 0.961$\pm$0.007* & 0.595$\pm$0.010* & 0.991$\pm$0.001* & 0.277$\pm$0.006 \\
EmbSMOTE & 0.872$\pm$0.003 & 0.911$\pm$0.002* & 0.860$\pm$0.008* & 0.958$\pm$0.004* & \textbf{0.597$\pm$0.007} & 0.989$\pm$0.001 & \textbf{0.292$\pm$0.001} \\
LLM-Paraphrase & \textbf{0.879$\pm$0.004} & 0.909$\pm$0.002 & 0.823$\pm$0.002 & 0.948$\pm$0.004 & 0.557$\pm$0.006 & 0.990$\pm$0.001* & 0.237$\pm$0.011 \\
LCG & 0.872$\pm$0.006* & 0.906$\pm$0.001 & 0.832$\pm$0.005 & 0.931$\pm$0.002 & 0.522$\pm$0.005 & 0.989$\pm$0.001 & 0.232$\pm$0.012 \\
AugGPT & 0.874$\pm$0.002* & 0.909$\pm$0.002 & 0.827$\pm$0.006 & 0.934$\pm$0.006 & 0.540$\pm$0.005 & 0.991$\pm$0.001* & 0.257$\pm$0.011 \\
CoTAM & 0.869$\pm$0.006 & 0.911$\pm$0.002* & 0.820$\pm$0.005 & 0.943$\pm$0.006 & 0.560$\pm$0.010 & \textbf{0.991$\pm$0.001} & 0.234$\pm$0.012 \\
LLM2LLM & 0.868$\pm$0.003 & 0.911$\pm$0.001* & 0.815$\pm$0.009 & 0.935$\pm$0.002 & 0.524$\pm$0.019 & 0.990$\pm$0.001* & 0.203$\pm$0.008 \\
CIEGAD & 0.869$\pm$0.005 & 0.904$\pm$0.006 & 0.821$\pm$0.011 & 0.955$\pm$0.007 & 0.521$\pm$0.017 & 0.988$\pm$0.002 & 0.229$\pm$0.030 \\
VoidGen      & 0.870$\pm$0.007* & 0.907$\pm$0.004 & 0.836$\pm$0.010 & \textbf{0.964$\pm$0.004} & 0.557$\pm$0.020 & 0.989$\pm$0.001 & 0.216$^{\dagger}$ \\
\bottomrule
\end{tabular}%
}
\end{table}

As shown in Table~\ref{tab:master-f1}, it can be observed that a consistent pattern emerges across all seven datasets.
In particular, the four classical methods (EDA, AEDA, BackTrans, EmbSMOTE) occupy the top positions on 6 of 7 datasets, and EmbSMOTE attains the best or statistically tied-best result on the imbalanced multi-class datasets (GoEmotions-13, GoEmotions-28, and Emo).
Importantly, none of the six LLM-based methods (LLM-Paraphrase, LCG, AugGPT, CoTAM, LLM2LLM, and CIEGAD) statistically matches the classical methods on these datasets; the gap reaches up to $\Delta\text{F1}_\text{macro}{\approx}0.063$ on GoEmotions-28 (IR${\approx}527.7$).

Stratifying the results by dataset characteristics reveals three distinct regimes.
First, on the balanced large-scale datasets, namely SST-2, AG~News, and DBpedia, macro F1 for nearly all methods concentrates within $\pm 0.005$, rendering the choice of augmentation method empirically irrelevant.
Second, on the imbalanced multi-class datasets, namely Emo, TREC, GoEmotions-13, and GoEmotions-28, classical methods consistently outperform LLM-based methods.
Third, the proposed VoidGen attains the best value of $0.964$ on TREC while slightly trailing EmbSMOTE on the imbalanced datasets.
Because macro F1 averages can conceal dynamics visible only at class resolution, we present a fine-grained per-class analysis in Section~\ref{sec:per-class} that exposes structural patterns that this aggregate metric cannot capture.

\subsection{Evaluation on Statistical Significance}\label{sec:welch}
To verify that the patterns observed in Section~\ref{sec:master-f1} are statistically significant, we conduct a systematic comparison using Welch's $t$-test.
Table~\ref{tab:welch} summarizes the win/tie/loss record of each method against EmbSMOTE using Welch's $t$-test at $p<0.05$ across all seven datasets.
For transparency, the per-cell Welch's $t$ statistic, Welch--Satterthwaite degrees of freedom, and $p$-value underlying this summary are reported in full in~\ref{app:welch}.
Each dataset is treated as an independent task, and no multiple-comparison correction is applied across datasets.
Our rationale for selecting EmbSMOTE as the reference is twofold.
First, EmbSMOTE is the de facto standard for imbalanced tabular classification and, as a representative retrieval-based oversampling method, represents the empirical ceiling of the classical family.
Second, as shown in Table~\ref{tab:master-f1}, EmbSMOTE ranks in the top group on 6 of 7 datasets and consistently achieves the best or tied-best performance under imbalanced multi-class settings; it therefore serves as a substantive strong baseline in this benchmark.

\begin{table}[t]
\centering
\caption{Welch's $t$-test against EmbSMOTE ($p<0.05$): win/tie/loss counts across the seven datasets. The N/A column counts cells in which the test was not computable. For VoidGen on GoEmotions-28, only $n=1$ seed completed (the other four failed with ZeroDivisionError), so Welch's test is not computable and the cell is recorded as N/A.}
\label{tab:welch}
\begin{tabular}{l c c c c}
\toprule
Method & Win & Tie & Loss & N/A \\
\midrule
No-Aug & 0 & 4 & 3 & 0 \\
EDA & 0 & 6 & 1 & 0 \\
AEDA & 0 & 7 & 0 & 0 \\
BackTrans & 0 & 6 & 1 & 0 \\
LLM-Paraphrase & 1 & 2 & 4 & 0 \\
LCG & 0 & 2 & 5 & 0 \\
AugGPT & 0 & 3 & 4 & 0 \\
CoTAM & 1 & 2 & 4 & 0 \\
LLM2LLM & 0 & 3 & 4 & 0 \\
CIEGAD & 0 & 4 & 3 & 0 \\
VoidGen      & 0 & 4 & 2 & 1 \\
\bottomrule
\end{tabular}
\end{table}

Stratifying the results by method family reveals three tiers that correspond to design philosophy.
The first tier comprises the four classical methods (EDA, AEDA, BackTrans, and EmbSMOTE), which remain within the statistical equivalence region of EmbSMOTE and suffer at most one loss each; surface-form-preserving perturbation and embedding-space retrieval reproducibly achieve parity with the reference in this benchmark.
The second tier consists of the LLM-based methods, namely LLM-Paraphrase, LCG, AugGPT, CoTAM, and LLM2LLM, which collectively record at least four losses each and only two wins in total, failing to significantly outperform classical retrieval regardless of prompt strategy or generation sophistication.
The third tier contains even the most geometrically sophisticated method, CIEGAD, which reaches only $0/4/3$ and thus approaches but does not surpass the classical tier without a single win.
Crucially, this three-tier structure does not arise from differences in method complexity or generative capability, but rather corresponds to whether class membership is structurally guaranteed by design, an observation that motivates the mechanistic explanation developed from Section~\ref{sec:per-class} onwards.

Particularly noteworthy is the No-Aug record of $0/4/3$.
All three losses for No-Aug occur on the imbalanced multi-class datasets, namely Emo, GoEmotions-13, and GoEmotions-28, which demonstrates that some form of augmentation does provide a statistically significant performance improvement in these settings.
By contrast, No-Aug is statistically equivalent to EmbSMOTE on SST-2, AG~News, DBpedia, and TREC, corroborating the existence of saturated regions in which augmentation is redundant.
Accordingly, the regime in which augmentation genuinely matters, as captured by this benchmark, is concentrated in the imbalanced multi-class setting, and it is precisely in this regime that LLM-based augmentation consistently lags behind classical methods across all three method families.

\subsection{Per-Class F1 Analysis}\label{sec:per-class}
\begin{figure}[t]
\centering
\makebox[\linewidth][c]{%
\includegraphics[width=1.4\linewidth]{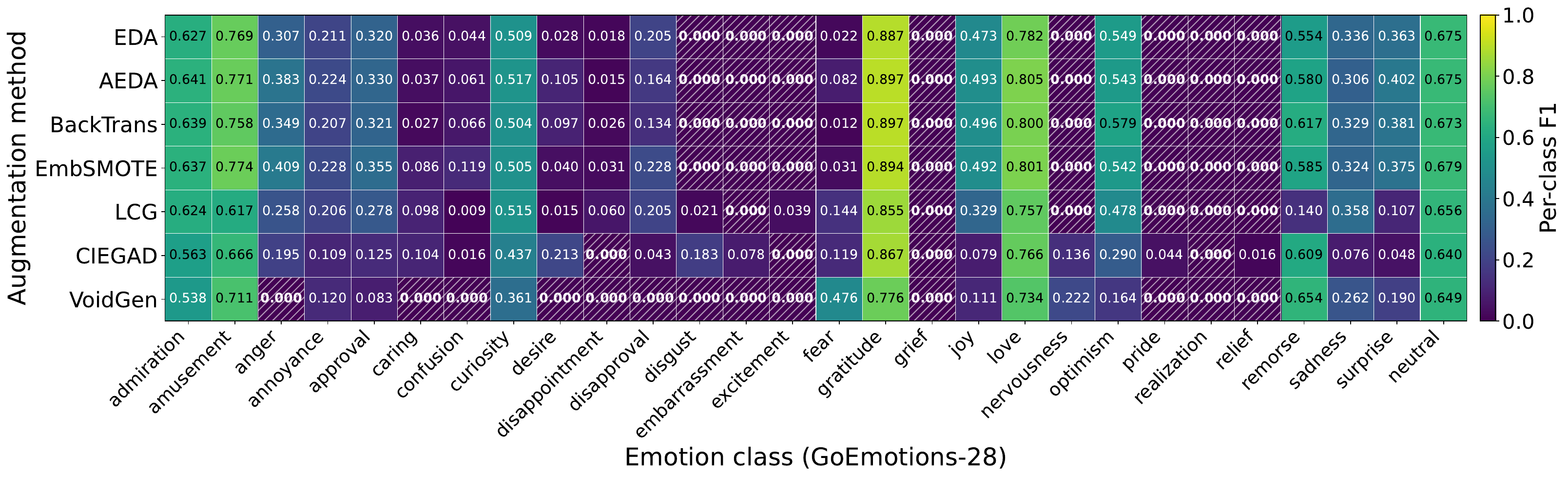}%
}
\caption{Per-class F1 heatmap for GoEmotions-28, showing eight representative methods (rows) and the 28 classes (columns).}
\label{fig:per-class-f1}
\end{figure}

The macro F1 ranking conceals what we regard as the most important benchmark findings of this study, which surface only at per-class resolution.
Figure~\ref{fig:per-class-f1} shows a per-class F1 heatmap for eight methods on GoEmotions-28, visually summarizing the dynamics in the three class regions described below.

On GoEmotions-28, six of the 28 classes, namely \textit{embarrassment}, \textit{grief}, \textit{nervousness}, \textit{pride}, \textit{realization}, and \textit{relief}, uniformly yield $\text{F1}{=}0.000$ across all five seeds for the four classical methods and LCG.
These six classes correspond to extremely rare tail classes in the $n_{\text{train}}=5{,}000$ split, and the resulting zero performance stems from a fundamental absence of training signal: EDA, AEDA, and BackTrans induce only minimal changes in representation space; EmbSMOTE faces an intrinsic diversity constraint when interpolating from only a handful of embeddings; and LCG cannot reach the semantic region of a class without sufficient anchors.
In short, none of these methods can rescue classes for which the training set provides essentially no signal. This constraint holds regardless of generation sophistication, and augmentation should accordingly be understood as a regularizer that presupposes a sufficient per-class signal.
By contrast, the generation-driven methods CIEGAD and VoidGen occasionally achieve non-zero F1 on one or more of these rare classes, with observed values ranging from $0.07$ to $0.33$.
This is attributable to class-conditioned generation, which can produce synthetic samples even for the rarest classes.
However, these sporadic non-zero values are too small in magnitude to alter the macro F1 ranking.
We note that the per-class predictions for LLM-Paraphrase and AugGPT on GoEmotions-28 were not available locally at the time of writing, so the tail-class behavior of those two methods remains unverified.

Class-size-stratified behavior is contrastive across three regions.
In the tail-class region ($<20$ examples), the four classical methods and LCG show $\text{F1}{=}0.000$ across all seeds, while only generative methods occasionally record non-zero values.
In the head-class region ($\geq250$ examples), comprising gratitude, love, amusement, neutral, and admiration, all methods uniformly reach per-class F1 between $0.62$ and $0.90$, with differences among the top three methods of less than $\pm 0.01$.
Augmentation strategy has no practical effect on classification performance where training examples are plentiful.
The middle-class region ($20$--$250$ examples) is where the core finding of this paper resides.
In this range, training signal exists but is sparse, and consequently augmentation quality translates directly into classification performance.
Specifically, EmbSMOTE and the other classical methods stably accumulate incremental signal by faithfully preserving the within-class distributional geometry, whereas LLM-based methods inject gradient noise through off-class generated samples and lose their competitive edge in this region.
These middle-class dynamics are the primary driver of the LLM-versus-classical macro F1 gap observed in Sections~\ref{sec:welch} and \ref{sec:llm-family}.
The advantage of class-structure preservation appears neither in tail classes nor in head classes, but is concentrated precisely in the middle classes where per-class signal exists yet is sparse.
This finding directly supports the central thesis of this paper, namely that ``class-structure preservation beats diversity,'' at the granularity of individual class behavior.

\subsection{Diversity--Performance Correlation Analysis}\label{sec:div-vs-pres}
The three-tier structure in Section~\ref{sec:welch} does not explain why classical methods outperform LLM-based methods.
In this section, we directly test the prevailing explanation for augmentation effectiveness, namely that augmentation diversity matters.
The common motivation for LLM-based augmentation is that LLMs generate more diverse text than retrieval- or rule-based methods, and that this diversity translates into stronger downstream classifiers.
To examine this hypothesis directly, for each (method, dataset) pair we measure five distributional properties of the augmented set $\widetilde{\mathcal{D}}$: uniqueness rate, TTR, mean character length, mean token count, and normalized label entropy; their Pearson correlations with the corresponding macro F1 are computed.
Table~\ref{tab:diversity-corr} summarizes both within-dataset Pearson $r$ averaged across datasets and pooled correlations across all (method, dataset) cells, with dataset-level F1 baseline differences controlled.

\begin{table}[t]
\centering
\caption{Pearson correlations between augmentation diversity metrics and macro F1, computed across the 11 augmentation methods.}
\label{tab:diversity-corr}
\begin{tabular}{l c c}
\toprule
Diversity metric & $r$ (within-dataset mean) & $r$ (pooled) \\
\midrule
Uniqueness rate         & $-0.166$ & $-0.031$ \\
TTR           & $+0.514$ & $+0.110$ \\
Mean character length                    & $-0.335$ & $+0.353$ \\
Normalized label entropy      & $-0.170$ & $+0.723$ \\
Mean token count                & $-0.335$ & $+0.317$ \\
\bottomrule
\end{tabular}
\end{table}

Three observations stand out.
First, surface-level diversity does not promote downstream performance.
The uniqueness rate of augmented sets exhibits at most a weakly negative correlation with macro F1 ($r{=}-0.166$).
EmbSMOTE shows the lowest uniqueness rate across datasets, between $58\%$ and $61\%$, yet attains the best or tied-best performance on 6 of 7 datasets, which empirically refutes the naive view that ``diversity is good.''
Second, only corpus-level vocabulary diversity exhibits a positive correlation with performance.
TTR shows a moderate positive correlation ($r{=}+0.514$; Fisher's combined $p{=}0.018$ across the seven within-dataset tests), which indicates that methods whose augmented corpora draw on a broader vocabulary tend to yield stronger downstream classifiers.
By contrast, the uniqueness rate is statistically negligible (combined $p{=}0.53$), whereas vocabulary breadth is reproducible across datasets, providing evidence that distinguishes which axis of diversity actually matters.
Third, the side effects of LLM generation are negatively associated with performance.
Mean character length and forced label uniformity both yield negative correlations with macro F1 ($r{=}-0.335$ and $r{=}-0.170$, respectively), and both are characteristic artifacts of LLM-based augmentation: chat-aligned LLMs produce longer texts than the original training corpus, while methods such as LCG and CIEGAD explicitly steer the label distribution toward uniformity to address class imbalance.
Our data suggest that both effects are net-harmful, providing a mechanistic explanation of the LLM-versus-classical gap reported in Section~\ref{sec:welch}.

\subsection{Augmented Text Visualization}\label{sec:viz}
To complement the quantitative analysis, two visualizations show the extent to which augmented texts deviate from the original training distribution.
In both figures, the original training data ($n_{\text{train}}=5000$, fixed sample at seed $=42$) is shown in the background, with the distribution of augmented texts from each method overlaid on top.

\begin{figure}[t]
\centering
\includegraphics[width=\linewidth]{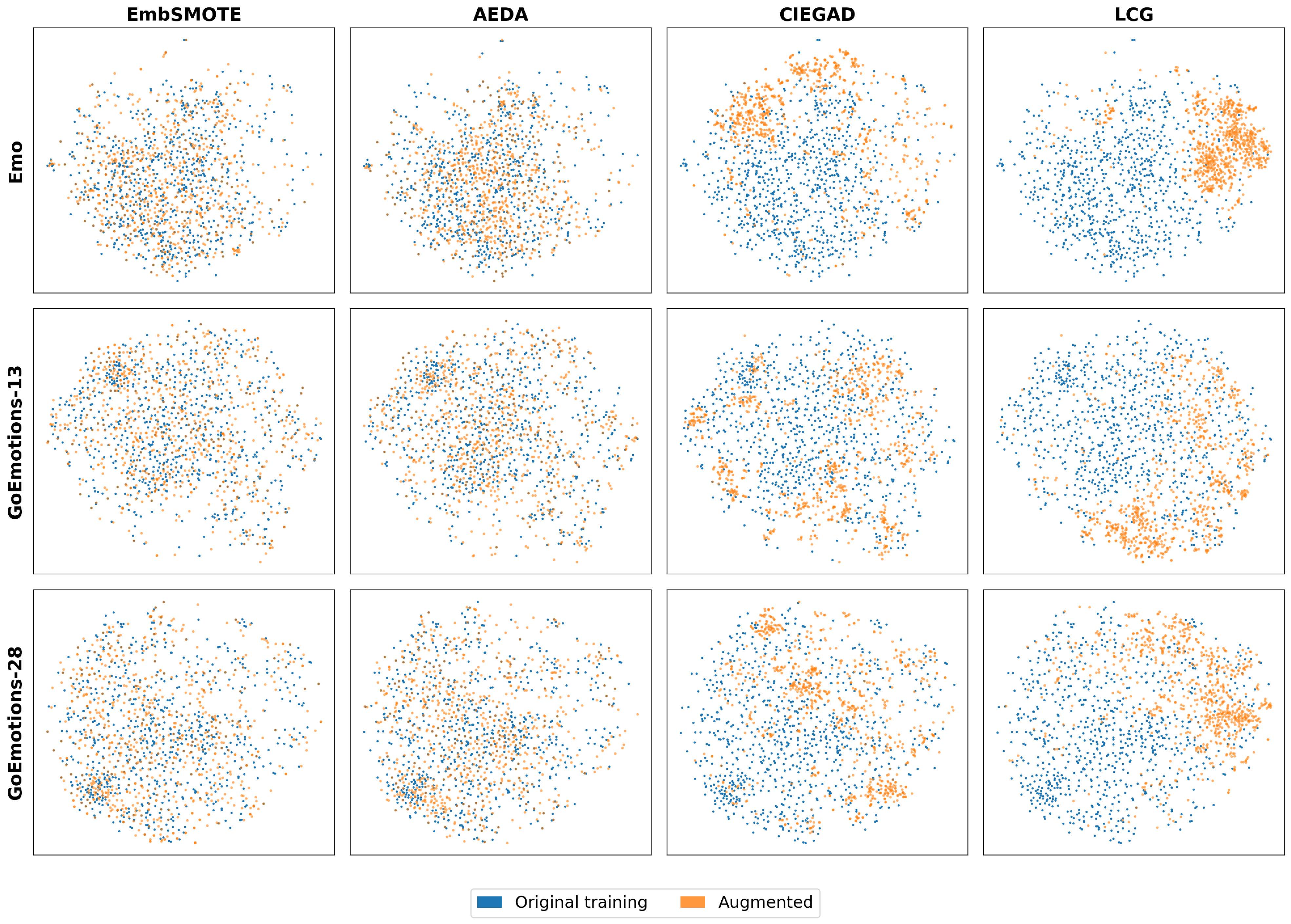}
\caption{SBERT embeddings of augmented texts (orange) and original training texts (blue) for four representative methods (columns: EmbSMOTE, AEDA, CIEGAD, LCG), co-projected into a shared t-SNE space.}
\label{fig:tsne}
\end{figure}

Figure~\ref{fig:tsne} shows SBERT embeddings of augmented texts (orange) and original training texts (blue) for four representative methods, namely EmbSMOTE, AEDA, CIEGAD, and LCG, co-projected into a shared t-SNE space on the three most characteristic imbalanced multi-class emotion datasets.
The figure is laid out as a $3\times 4$ grid, in which rows correspond to datasets and columns to methods.
As can be observed, the augmented point clouds of EmbSMOTE and AEDA overlap almost entirely with the blue point cloud of the training distribution in both density and spatial extent, rendering them visually indistinguishable in the embedding space.
This is expected by design: EmbSMOTE obtains synthetic samples by nearest-neighbor retrieval of real training texts, and AEDA performs only minor surface-form perturbation, and both methods therefore structurally preserve the geometry of the training distribution.

By contrast, the augmented points generated by CIEGAD form several dense local clusters in regions distant from the periphery of the training distribution, concentrating in areas where training points are sparse.
Particularly on GoEmotions-28, multiple dense clusters appear near the margins of the training manifold, which reflects the behavior of CIEGAD's inverse-frequency cluster budget allocation that directs augmentation resources toward rare-class target regions.
LCG exhibits an even more pronounced pattern, forming one or two extremely dense clusters in regions outside the training distribution across all datasets; this suggests mode-collapse-like behavior in which generation collapses to specific class- or style-prototypes, a consequence of label-conditioned generation operating without anchors, which causes the LLM to repeatedly emit similar expression patterns.
These off-manifold point clouds do not match the natural class-wise distribution width and therefore fail to provide an effective signal for rescuing rare classes, which is consistent with the F1 deficits reported in Section~\ref{sec:welch}.

\begin{figure}[t!]
\centering
\includegraphics[height=0.87\textheight,keepaspectratio]{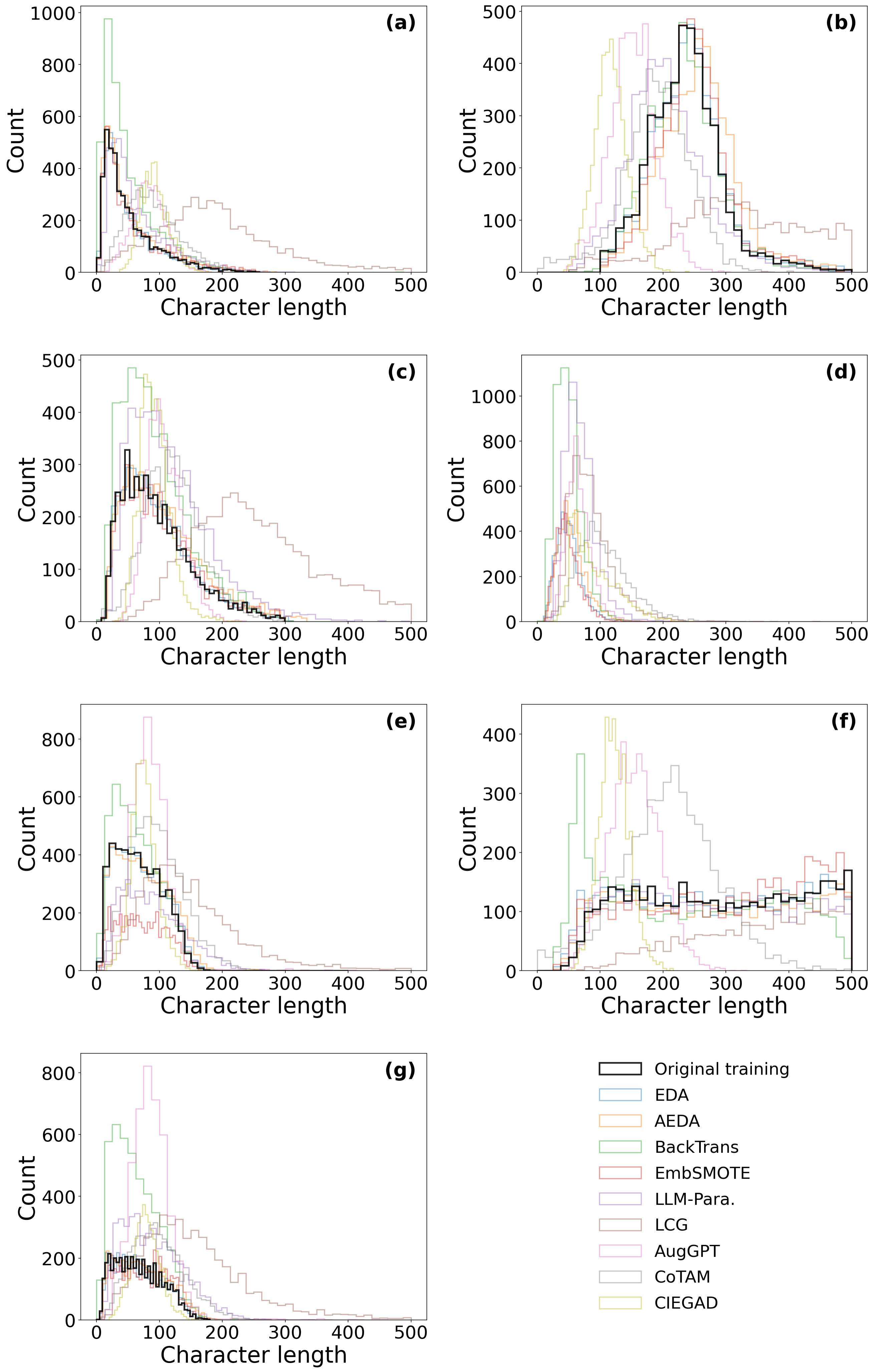}
\caption{Character-length distributions of the original training texts (black step histogram) and of the augmented texts for each method, across all seven datasets (rows).}
\label{fig:charlen}
\end{figure}

Figure~\ref{fig:charlen} overlays the character-length distributions of each augmentation method on the original training-text distribution (black step histogram) across all seven datasets.
As can be observed, the classical and retrieval-based methods, namely EDA, AEDA, EmbSMOTE, and BackTrans, largely preserve shapes consistent with the training distribution; in particular, EmbSMOTE overlaps perfectly because it reuses real training texts.
By contrast, the LLM-based methods display two qualitatively distinct behaviors.
First, LCG systematically shifts the distribution toward much longer texts across all datasets, with mean character lengths reaching approximately two to four times those of the training distribution.
Second, LLM-Paraphrase, AugGPT, and CoTAM exhibit a non-uniform directionality: they shift toward longer texts on short-text datasets (SST-2, Emo, GoEmotions-13, and GoEmotions-28) while shifting toward shorter texts on long-text datasets (AG~News and DBpedia).
This indicates that these methods do not preserve the original text-length characteristics, but instead aggregate their outputs toward a fixed intermediate length band regardless of the dataset.
In all cases, LLM-based augmentation fails to preserve the length properties of the training distribution, which provides a mechanistic explanation for the quantitative finding that mean character length is negatively correlated with macro F1 (see Section~\ref{sec:div-vs-pres}).

\subsection{LLM Family Sensitivity Analysis}\label{sec:llm-family}
A natural counterargument to the negative results of Section~\ref{sec:welch} is that Llama-3.1-8B is simply too weak.
If artifacts specific to a single LLM family were the primary driver of the gap, replacing the LLM with a stronger model should reverse the ranking.
To rule out this possibility, we conduct a replication experiment with a different LLM family in this section.
To disentangle the ``LLM is the bottleneck'' hypothesis from the ``generate-then-verify formulation is the bottleneck'' hypothesis, we re-run CIEGAD with both Llama-3.1-8B-Instruct and Qwen3-8B under identical algorithmic hyperparameters, prompts, and judge thresholds.
Both LLMs are of the same 8B class, and we disable the chain-of-thought thinking mode of Qwen3-8B so that both backends generate single-pass completions.
Results over five seeds are presented in Table~\ref{tab:llm-family}.

\begin{table}[t]
\centering
\caption{LLM-family sensitivity: CIEGAD with Llama-3.1-8B vs.\ Qwen3-8B (mean $\pm$ standard deviation over five seeds). $\Delta$ denotes the Qwen $-$ Llama difference in macro F1.}
\label{tab:llm-family}
\begin{tabular}{l c c c}
\toprule
Dataset & CIEGAD & CIEGAD-Qwen & $\Delta$ \\
\midrule
SST-2 & 0.869$\pm$0.005 & 0.867$\pm$0.004 & $-0.002$ \\
AG~News & 0.904$\pm$0.006 & 0.906$\pm$0.005 & $+0.001$ \\
Emo & 0.821$\pm$0.011 & 0.826$\pm$0.011 & $+0.005$ \\
TREC & 0.955$\pm$0.007 & 0.946$\pm$0.005 & $-0.009$ \\
GoEmotions-13 & 0.521$\pm$0.017 & 0.545$\pm$0.010 & $+0.024$ \\
DBpedia & 0.988$\pm$0.002 & 0.987$\pm$0.002 & $-0.001$ \\
GoEmotions-28 & 0.229$\pm$0.030 & 0.281$\pm$0.022 & $+0.052$ \\
\bottomrule
\end{tabular}
\end{table}

Results divide clearly along the imbalance axis introduced in Section~\ref{sec:welch}.
In the saturated settings, namely SST-2, AG~News, DBpedia, and TREC, the two LLM families are within $\pm 0.01$ macro F1 on all datasets, rendering the choice of LLM empirically irrelevant; both backends trail EmbSMOTE by similarly small margins.
In the imbalanced multi-class emotion settings, however, a more nuanced picture emerges that is structured by the degree of imbalance.
On Emo ($K{=}6$, IR${\approx}9.49$), the two LLM families are indistinguishable ($\Delta{=}{+}0.005$, Welch's $t{=}0.70$, $\mathrm{df}{=}8.0$, $p{=}0.50$).
On GoEmotions-13 ($K{=}13$, IR${\approx}12.4$), Qwen3 is significantly stronger ($\Delta{=}{+}0.024$, $t{=}2.74$, $\mathrm{df}{=}6.6$, $p{=}0.031$).
On the most fine-grained and most imbalanced dataset, GoEmotions-28 ($K{=}28$, IR${\approx}527.7$), the gap is largest and most significant ($\Delta{=}{+}0.052$, $t{=}3.09$, $\mathrm{df}{=}7.4$, $p{=}0.016$).
Accordingly, the advantage of Qwen3 grows monotonically with increasing class imbalance, which we interpret as evidence that the newer instruction-tuned LLM is better at maintaining on-class generation under tail-class pressure, a regime in which the prompt-following of Llama-3.1 degrades.

Crucially, however, upgrading the LLM does not close the gap to retrieval-based EmbSMOTE.
Even at its best, CIEGAD-Qwen still falls below EmbSMOTE on all imbalanced multi-class datasets, and the win/tie/loss summary against EmbSMOTE is essentially unchanged between the two LLM families.
In other words, the bottleneck of LLM-based augmentation is not LLM quality at the 8B scale, but rather the absence of class-structure preservation in the generate-then-verify formulation.

\subsection{Training Set Size Sensitivity Analysis}\label{sec:n-train-curve}
Our benchmark adopts $n_{\text{train}}=5{,}000$ as the standard setting, but this scale may favor classical methods.
To examine the widely shared hypothesis that ``LLM augmentation is most effective when training data are scarce,'' we conduct a sensitivity analysis in which the training-set size is varied systematically.
We sweep $n_{\text{train}}\in\{500,\,1000,\,2000,\,5000\}$ on GoEmotions-13, keeping the classifier and augmentation pipelines identical to those of Section~\ref{sec:experiments}, in order to examine the behavior of augmentation as the training set shrinks.
To span the spectrum, we compare four methods: EmbSMOTE and AEDA, LLM-Paraphrase, and CIEGAD.
All cells are run with five seeds, and the means and standard deviations are reported in Table~\ref{tab:n-train-curve}.

\begin{table}[t]
\centering
\caption{Macro F1 as a function of $n_{\text{train}}$ on GoEmotions-13 (mean $\pm$ standard deviation over five seeds). Bold indicates the best method at each value of $n_{\text{train}}$.}
\label{tab:n-train-curve}
\begin{tabular}{r c c c c }
\toprule
$n_{\text{train}}$ & EmbSMOTE & AEDA & CIEGAD & LLM-Paraphrase \\
\midrule
500 & $0.074\pm0.023$ & $0.059\pm0.018$ & $\bm{0.079\pm0.028}$ & $0.047\pm0.002$ \\
1{,}000 & $0.184\pm0.024$ & $0.147\pm0.016$ & $\bm{0.200\pm0.025}$ & $0.134\pm0.009$ \\
2{,}000 & $\bm{0.394\pm0.004}$ & $0.382\pm0.009$ & $0.380\pm0.016$ & $0.262\pm0.018$ \\
5{,}000 & $\bm{0.597\pm0.007}$ & $0.596\pm0.010$ & $0.521\pm0.017$ & $0.560\pm0.008$ \\
\bottomrule
\end{tabular}
\end{table}

Contrary to the prevailing intuition that ``the LLM-versus-classical gap widens as data shrinks,'' two clear patterns emerge.
First, the method ranking is not constant along the curve: it reverses at approximately $n_{\text{train}}{\approx}2{,}000$.
For $n_{\text{train}}{\le}1{,}000$, CIEGAD achieves the best performance, slightly outperforming EmbSMOTE.
In deeply low-resource regimes, the genuinely novel content generated by LLM-based augmentation provides a small but measurable advantage that classical perturbation methods cannot match.
By contrast, for $n_{\text{train}}{\ge}2{,}000$, the order reverses: EmbSMOTE and AEDA take the lead, and the gap with CIEGAD widens as $n_{\text{train}}$ grows, reaching $\Delta{=}{+}0.076$ macro F1 at $n_{\text{train}}{=}5{,}000$.
Second, the anchor-based LLM-Paraphrase is inferior at all values of $n_{\text{train}}$.
Unlike CIEGAD, which can introduce class-conditioned content, LLM-Paraphrase is constrained to remain near each anchor sentence, contributing little beyond what classical character-level perturbation already provides.

The qualitative implication is that the value of LLM-based augmentation is regime-dependent rather than monotonically related to the LLM-versus-classical gap.
LLM augmentation is most useful when training data are genuinely scarce ($n_{\text{train}}{\le}1{,}000$), and it offers no benefit, in fact underperforming classical alternatives, once at least several thousand training examples per class are available.
This finding refines rather than contradicts our main F1 results.
The $n_{\text{train}}{=}5{,}000$ setting employed throughout the rest of the benchmark is a regime in which classical retrieval-based oversampling is empirically optimal, and the LLM-versus-classical gap identified there should most safely be interpreted conditional on sufficient training data being available.

\section{Discussion}\label{sec:discussion}
\subsection{Why Classical Methods Win}\label{sec:why-classical}
The results of Section~\ref{sec:results} converge on a single mechanistic explanation, which is best understood as a question of structural class-membership guarantees.
The reason that retrieval-based EmbSMOTE is statistically equivalent to or superior to all LLM-based competitors reduces to whether the class membership of augmented examples is structurally guaranteed by design.
EmbSMOTE returns existing within-class training texts, so every $(\tilde{x},\tilde{y})$ in $\widetilde{\mathcal{D}}$ is by definition on-distribution and correctly labeled.
By contrast, LLM-based methods rely on prompt conditioning and post-hoc filtering to ensure class membership, and whenever generation drifts, label noise is injected into the gradient that pulls the classifier away from the true decision boundary.
The t-SNE visualizations in Section~\ref{sec:viz}, where EmbSMOTE and AEDA overlap with the training distribution while CIEGAD clusters at the manifold periphery and LCG forms off-distribution blobs, provide direct visual corroboration of this off-class gradient source.

This view is consistent with both the diversity analysis (Section~\ref{sec:div-vs-pres}) and the per-class analysis (Section~\ref{sec:per-class}).
Taken together, these analyses indicate that surface-level diversity (uniqueness rate) barely predicts macro F1; in fact, EmbSMOTE achieves the lowest uniqueness rate in this benchmark yet attains best-class performance.
This demonstrates that augmentation functions as an effective regularizer not as a source of new information, but only insofar as it does not disturb the class-conditioned geometry.
Furthermore, the gap between LLM and classical methods is concentrated not in tail classes nor in head classes but in the middle-class region ($20$--$250$ examples), where training signal exists yet is sparse.
In that region, classical methods stably accumulate incremental signal by faithfully preserving within-class distributional geometry, whereas LLM-based methods contaminate gradient updates with off-boundary generations and lose their competitive edge.
This finding suggests that the main thesis of this paper, namely that ``class-structure preservation beats diversity,'' is substantiated at the mechanistic level rather than as a mere empirical regularity.

\subsection{The LLM Quality Bottleneck}\label{sec:llm-bottleneck}
A natural counterargument to these negative findings is that Llama-3.1-8B is simply too weak to support fair comparison.
In Section~\ref{sec:llm-family}, the most sophisticated method, CIEGAD, was replicated with the more recent Qwen3-8B to rule out this possibility.
Results are clear: LLM family differences appear only in high-imbalance settings (maximum $\Delta\text{F1}{\approx}0.052$ on GoEmotions-28), whereas for saturated tasks the choice of LLM makes no discernible difference, and both trail EmbSMOTE by similarly small margins.
This is consistent with the interpretation that newer instruction-tuned LLMs are better able to keep class-conditioned generation on-class under tail-class pressure.

However, what is decisive is that even a stronger LLM cannot close the gap to EmbSMOTE in principle.
Compared with CIEGAD, CIEGAD-Qwen still falls significantly below EmbSMOTE on imbalanced multi-class datasets except GoEmotions-28.
On GoEmotions-28 specifically, Qwen3 narrows the gap to statistical equivalence ($p{=}0.32$), but this means that an entire generation of LLM progress was required to match a level that retrieval-based oversampling achieves with zero parameter updates.
The approximately 80\% gap reduction on GoEmotions-28 ($\Delta{\approx}0.063\to0.011$) is consistent with the possibility that pure scaling may eventually close the residual, while simultaneously highlighting that the same result is obtainable via class-structure preservation at no additional cost.
Taken together, these results indicate that the cause of this gap is not LLM weakness per se.
Rather, the fact that upgrading the LLM family and scale does not close the gap suggests that the bottleneck resides in the structural constraint of the generate-then-verify paradigm, namely its inability to guarantee the class membership of augmented examples.

\subsection{Task-Dependent Method Selection}\label{sec:task-dep}
The benchmark naturally divides into two environments with substantially different practical implications, from which method-selection rules are derived.
First, in saturated settings where the unaugmented macro F1 exceeds $0.85$ on binary or low-class-count tasks, the choice of augmentation method is empirically irrelevant, and the cheapest classical option suffices.
The $+10.52\%$ gain from LLM paraphrase reported by Wang et al.~\cite{wang2025diversity} on balanced binary benchmarks is also consistent with this interpretation, suggesting that improvements in this regime represent noise near the performance ceiling rather than a systematic advantage.
Second, for fine-grained imbalanced multi-class classification, EmbSMOTE should be adopted as the default, and LLM-based augmentation should be required to satisfy a high evidentiary standard before deployment.
It is in this regime that method differences are largest (up to $\Delta\text{F1}{=}0.13$ on GoEmotions-28) and rankings are consistent.
These rules directly contradict the dominant assumption that more sophisticated generative augmentation is universally superior, and demonstrate that the value of augmentation depends decisively on the structure of the classification problem itself.

\subsection{Limitations}\label{sec:limits}
Several limitations of this study should be noted.
First, the LLMs evaluated are limited to two models of the 8B class (Llama-3.1-8B and Qwen3-8B), and conclusions may change with 70B-class or GPT-4-class models.
Second, all seven datasets are English text classification tasks; domain-specific and multilingual settings are out of scope, and this study intentionally focuses on imbalanced multi-class settings where the retrieval--generation gap is most pronounced.
Third, the classifier is fixed to a frozen DistilBERT, and it is unverified whether rankings are preserved with larger encoders; however, prior benchmarks~\cite{chen2023empirical} suggest that the interaction between classifier capacity and augmentation strategy is small.
Fourth, the augmentation ratio is fixed at $r{=}1.0$, although the $n_{\text{train}}$ analysis (Section~\ref{sec:n-train-curve}) partially compensates for this design choice.
Finally, on the statistical side, Welch's $t$-test behaves conservatively under non-normal distributions.

Two method-specific limitations also apply.
First, there is a generation-budget asymmetry: CIEGAD-family methods select the best from three candidates per anchor, whereas VoidGen uses a single candidate per target.
Increasing VoidGen's budget would likely bring its performance closer to CIEGAD-Qwen, but the LLM-quality ceiling demonstrated in Section~\ref{sec:llm-bottleneck} suggests that it would not surpass that ceiling.
Furthermore, VoidGen failed with a ZeroDivisionError on 4 of 5 seeds for GoEmotions-28, owing to numerical instability in the void detection procedure (division by zero under extreme tail-class sparsity); the reported value of $0.216$ therefore reflects a single seed and is treated as N/A in Table~\ref{tab:welch}.
Importantly, this failure does not affect the central claims; if anything, it reinforces the conclusion that even the most sophisticated targeting cannot rescue the rarest classes.
Nevertheless, numerical stabilization of the void detection procedure remains a future implementation priority.

\section{Conclusion}\label{sec:conclusion}
In this study, we constructed an empirical NLP augmentation benchmark that includes EmbSMOTE as a strong reference method.
The central finding is a clear negative result against the premise that LLM-based augmentation is universally superior: all LLM-based augmentation methods evaluated were statistically equivalent or inferior to EmbSMOTE, and the gap widened as class imbalance increased.
From this, task-dependent decision rules were derived: in saturated binary and low-class-count tasks, the choice of augmentation method is practically irrelevant; in imbalanced multi-class settings, retrieval-based oversampling should be adopted as the default, and LLM-based augmentation should be required to meet a high evidentiary standard.

Three independent lines of evidence, namely correlation analysis of distributional metrics of augmented sets, per-class F1 analysis, and replication experiments with Llama-3.1 and Qwen3, established that the operative variable is not surface-level diversity (uniqueness rate does not predict macro F1) but class-conditional structural fidelity.
VoidGen, introduced as a methodological probe, performs comparably to EmbSMOTE on most datasets yet trails on the most imbalanced GoEmotions-28, reinforcing the conclusion that pre-generation void targeting alone is insufficient to guarantee structural fidelity and that class-structure preservation is the more fundamental requirement.

In future work, we will explore three directions.
First, we will verify whether scaling to 70B-class or GPT-4-class models closes the gap identified in this work.
Second, we will investigate hybrid methods that combine structural preservation through retrieval with vocabulary diversity through rewriting.
Third, we will extend the benchmark to multilingual, multimodal, and structured-output tasks to confirm whether the claim of structural fidelity holds beyond the settings of this study.

\appendix
\section*{Appendix}
\section{Per-Dataset, Per-Seed F1 Tables}\label{app:full-f1}
For full reproducibility, Table~\ref{tab:full-f1} reports the raw per-seed macro F1 for every (method, dataset) cell.
This is the input data for all aggregated statistics in Section~\ref{sec:results}.

\begingroup
\small
\begin{longtable}{l l c c c c c}
\caption{Per-seed macro F1 for all (method, dataset) cells. Seeds = \{42, 123, 456, 789, 1234\}.}
\label{tab:full-f1}\\
\toprule
Method & Dataset & s=42 & s=123 & s=456 & s=789 & s=1234 \\
\midrule
No-Aug & sst2 & 0.873 & 0.870 & 0.878 & 0.877 & 0.874 \\
 & ag\_news & 0.912 & 0.908 & 0.912 & 0.908 & 0.908 \\
 & emo & 0.782 & 0.809 & 0.818 & 0.793 & 0.803 \\
 & trec & 0.934 & 0.923 & 0.918 & 0.936 & 0.802 \\
 & go\_emotions & 0.455 & 0.474 & 0.462 & 0.449 & 0.462 \\
 & dbpedia & 0.989 & 0.993 & 0.990 & 0.989 & 0.991 \\
 & go\_emotions-28 & 0.173 & 0.172 & 0.183 & 0.200 & 0.182 \\
\midrule
EDA & sst2 & 0.876 & 0.882 & 0.882 & 0.878 & 0.866 \\
 & ag\_news & 0.909 & 0.909 & 0.912 & 0.909 & 0.910 \\
 & emo & 0.857 & 0.860 & 0.862 & 0.868 & 0.858 \\
 & trec & 0.960 & 0.956 & 0.942 & 0.953 & 0.965 \\
 & go\_emotions & 0.589 & 0.594 & 0.592 & 0.594 & 0.579 \\
 & dbpedia & 0.989 & 0.992 & 0.989 & 0.989 & 0.992 \\
 & go\_emotions-28 & 0.271 & 0.281 & 0.276 & 0.272 & 0.284 \\
\midrule
AEDA & sst2 & 0.869 & 0.869 & 0.880 & 0.871 & 0.869 \\
 & ag\_news & 0.912 & 0.909 & 0.912 & 0.912 & 0.911 \\
 & emo & 0.874 & 0.866 & 0.869 & 0.863 & 0.876 \\
 & trec & 0.956 & 0.962 & 0.943 & 0.957 & 0.950 \\
 & go\_emotions & 0.598 & 0.613 & 0.594 & 0.591 & 0.583 \\
 & dbpedia & 0.989 & 0.990 & 0.990 & 0.990 & 0.991 \\
 & go\_emotions-28 & 0.286 & 0.274 & 0.291 & 0.279 & 0.289 \\
\midrule
BackTrans & sst2 & 0.877 & 0.870 & 0.884 & 0.879 & 0.868 \\
 & ag\_news & 0.914 & 0.915 & 0.915 & 0.911 & 0.913 \\
 & emo & 0.854 & 0.855 & 0.847 & 0.840 & 0.857 \\
 & trec & 0.966 & 0.966 & 0.946 & 0.962 & 0.964 \\
 & go\_emotions & 0.604 & 0.604 & 0.600 & 0.584 & 0.581 \\
 & dbpedia & 0.989 & 0.992 & 0.991 & 0.992 & 0.989 \\
 & go\_emotions-28 & 0.278 & 0.268 & 0.282 & 0.273 & 0.282 \\
\midrule
EmbSMOTE & sst2 & 0.871 & 0.874 & 0.875 & 0.866 & 0.873 \\
 & ag\_news & 0.911 & 0.915 & 0.910 & 0.910 & 0.909 \\
 & emo & 0.864 & 0.854 & 0.860 & 0.849 & 0.871 \\
 & trec & 0.957 & 0.964 & 0.951 & 0.961 & 0.958 \\
 & go\_emotions & 0.595 & 0.605 & 0.601 & 0.599 & 0.584 \\
 & dbpedia & 0.989 & 0.990 & 0.988 & 0.989 & 0.990 \\
 & go\_emotions-28 & 0.294 & 0.290 & 0.292 & 0.293 & 0.292 \\
\midrule
LLM-Para. & sst2 & 0.881 & 0.884 & 0.877 & 0.882 & 0.871 \\
 & ag\_news & 0.911 & 0.906 & 0.912 & 0.909 & 0.909 \\
 & emo & 0.826 & 0.825 & 0.824 & 0.819 & 0.824 \\
 & trec & 0.948 & 0.954 & 0.947 & 0.943 & 0.949 \\
 & go\_emotions & 0.556 & 0.562 & 0.565 & 0.558 & 0.546 \\
 & dbpedia & 0.991 & 0.991 & 0.989 & 0.989 & 0.990 \\
 & go\_emotions-28 & 0.222 & 0.227 & 0.250 & 0.238 & 0.249 \\
\midrule
LCG & sst2 & 0.865 & 0.879 & 0.876 & 0.876 & 0.867 \\
 & ag\_news & 0.908 & 0.906 & 0.905 & 0.907 & 0.905 \\
 & emo & 0.834 & 0.839 & 0.828 & 0.826 & 0.836 \\
 & trec & 0.933 & 0.934 & 0.929 & 0.930 & 0.930 \\
 & go\_emotions & 0.521 & 0.525 & 0.525 & 0.525 & 0.511 \\
 & dbpedia & 0.990 & 0.990 & 0.989 & 0.991 & 0.988 \\
 & go\_emotions-28 & 0.226 & 0.232 & 0.232 & 0.217 & 0.254 \\
\midrule
AugGPT & sst2 & 0.875 & 0.871 & 0.877 & 0.874 & 0.870 \\
 & ag\_news & 0.913 & 0.909 & 0.908 & 0.908 & 0.907 \\
 & emo & 0.827 & 0.839 & 0.823 & 0.821 & 0.826 \\
 & trec & 0.936 & 0.924 & 0.931 & 0.939 & 0.940 \\
 & go\_emotions & 0.531 & 0.543 & 0.546 & 0.542 & 0.536 \\
 & dbpedia & 0.990 & 0.992 & 0.989 & 0.991 & 0.991 \\
 & go\_emotions-28 & 0.248 & 0.254 & 0.271 & 0.243 & 0.270 \\
\midrule
CoTAM & sst2 & 0.869 & 0.873 & 0.856 & 0.873 & 0.873 \\
 & ag\_news & 0.907 & 0.912 & 0.910 & 0.914 & 0.911 \\
 & emo & 0.816 & 0.818 & 0.814 & 0.827 & 0.824 \\
 & trec & 0.952 & 0.937 & 0.939 & 0.949 & 0.937 \\
 & go\_emotions & 0.558 & 0.569 & 0.574 & 0.550 & 0.547 \\
 & dbpedia & 0.991 & 0.990 & 0.991 & 0.991 & 0.992 \\
 & go\_emotions-28 & 0.228 & 0.234 & 0.228 & 0.223 & 0.258 \\
\midrule
LLM2LLM & sst2 & 0.863 & 0.869 & 0.870 & 0.870 & 0.866 \\
 & ag\_news & 0.910 & 0.912 & 0.913 & 0.912 & 0.911 \\
 & emo & 0.813 & 0.810 & 0.816 & 0.806 & 0.832 \\
 & trec & 0.935 & 0.937 & 0.936 & 0.933 & 0.935 \\
 & go\_emotions & 0.542 & 0.546 & 0.521 & 0.493 & 0.518 \\
 & dbpedia & 0.989 & 0.991 & 0.991 & 0.988 & 0.991 \\
 & go\_emotions-28 & 0.189 & 0.203 & 0.204 & 0.215 & 0.202 \\
\midrule
CIEGAD-Llama & sst2 & 0.872 & 0.869 & 0.876 & 0.862 & 0.868 \\
 & ag\_news & 0.911 & 0.898 & 0.909 & 0.897 & 0.906 \\
 & emo & 0.813 & 0.813 & 0.818 & 0.819 & 0.840 \\
 & trec & 0.955 & 0.958 & 0.961 & 0.943 & 0.959 \\
 & go\_emotions & 0.525 & 0.512 & 0.535 & 0.497 & 0.536 \\
 & dbpedia & 0.990 & 0.990 & 0.987 & 0.987 & 0.987 \\
 & go\_emotions-28 & 0.206 & 0.197 & 0.260 & 0.261 & 0.224 \\
\midrule
CIEGAD-Qwen & sst2 & 0.867 & 0.871 & 0.871 & 0.867 & 0.862 \\
 & ag\_news & 0.907 & 0.900 & 0.909 & 0.900 & 0.911 \\
 & emo & 0.818 & 0.825 & 0.823 & 0.817 & 0.845 \\
 & trec & 0.951 & 0.952 & 0.946 & 0.942 & 0.941 \\
 & go\_emotions & 0.555 & 0.539 & 0.541 & 0.533 & 0.556 \\
 & dbpedia & 0.990 & 0.988 & 0.987 & 0.985 & 0.986 \\
 & go\_emotions-28 & 0.274 & 0.267 & 0.314 & 0.293 & 0.258 \\
\midrule
VoidGen     & sst2 & 0.873 & 0.878 & 0.865 & 0.861 & 0.875 \\
 & ag\_news & 0.910 & 0.903 & 0.910 & 0.906 & 0.903 \\
 & emo & 0.832 & 0.848 & 0.845 & 0.830 & 0.824 \\
 & trec & 0.965 & 0.961 & 0.959 & 0.968 & 0.968 \\
 & go\_emotions & 0.588 & 0.549 & 0.549 & 0.563 & 0.536 \\
 & dbpedia & 0.991 & 0.989 & 0.988 & 0.988 & 0.989 \\
 & go\_emotions-28 & --- & --- & --- & 0.216 & --- \\
\bottomrule
\end{longtable}
\endgroup

\section{Detailed Welch's $t$-test Statistics}\label{app:welch}
Table~\ref{tab:welch-full} reports the per-cell Welch's $t$-test statistic, Welch--Satterthwaite degrees of freedom, and $p$-value for every (method, dataset) comparison against EmbSMOTE underlying the win/tie/loss summary in Section~\ref{sec:welch}. An asterisk denotes statistical significance at $p<0.05$. A negative $t$ indicates the method underperforms EmbSMOTE on average; a positive $t$ indicates the opposite. Cells reported as N/A correspond to VoidGen on GoEmotions-28, for which only one of five seeds produced a finite result (see Section~\ref{sec:limits}).

\begin{table}[t]
\centering
\caption{Per-cell Welch's $t$, Welch--Satterthwaite degrees of freedom, and $p$-value for each method against EmbSMOTE across seven datasets. Each cell reports the three quantities on three lines. $^{*}$ denotes $p<0.05$.}
\label{tab:welch-full}
\scriptsize
\setlength{\tabcolsep}{3pt}
\renewcommand{\arraystretch}{1.1}
\begin{tabular}{l c c c c c c c}
\toprule
Method & SST-2 & AG~News & Emo & TREC & GoEm-13 & DBpedia & GoEm-28 \\
\midrule
No-Aug & \makecell{$+1.22$\\$8.0$\\$0.256$} & \makecell{$-1.06$\\$8.0$\\$0.320$} & \makecell{$-7.96$\\$6.7$\\$0.000^{*}$} & \makecell{$-2.18$\\$4.1$\\$0.094$} & \makecell{$-24.71$\\$7.8$\\$0.000^{*}$} & \makecell{$+1.49$\\$6.3$\\$0.185$} & \makecell{$-21.63$\\$4.1$\\$0.000^{*}$} \\
EDA & \makecell{$+1.41$\\$5.9$\\$0.208$} & \makecell{$-1.01$\\$5.9$\\$0.353$} & \makecell{$+0.25$\\$5.8$\\$0.810$} & \makecell{$-0.66$\\$6.1$\\$0.531$} & \makecell{$-1.52$\\$7.6$\\$0.170$} & \makecell{$+1.32$\\$6.0$\\$0.233$} & \makecell{$-5.96$\\$4.6$\\$0.003^{*}$} \\
AEDA & \makecell{$-0.15$\\$6.9$\\$0.882$} & \makecell{$-0.07$\\$6.9$\\$0.947$} & \makecell{$+2.10$\\$6.7$\\$0.075$} & \makecell{$-1.20$\\$6.7$\\$0.270$} & \makecell{$-0.16$\\$7.2$\\$0.876$} & \makecell{$+2.30$\\$7.4$\\$0.053$} & \makecell{$-2.54$\\$4.3$\\$0.059$} \\
BackTrans & \makecell{$+1.14$\\$5.9$\\$0.298$} & \makecell{$+2.04$\\$7.4$\\$0.078$} & \makecell{$-1.89$\\$7.6$\\$0.097$} & \makecell{$+0.57$\\$6.3$\\$0.586$} & \makecell{$-0.32$\\$7.2$\\$0.761$} & \makecell{$+1.91$\\$6.6$\\$0.101$} & \makecell{$-5.32$\\$4.4$\\$0.004^{*}$} \\
LLM-Paraphrase & \makecell{$+2.71$\\$7.0$\\$0.030^{*}$} & \makecell{$-1.25$\\$8.0$\\$0.248$} & \makecell{$-8.90$\\$4.8$\\$0.000^{*}$} & \makecell{$-3.68$\\$7.9$\\$0.006^{*}$} & \makecell{$-8.24$\\$7.9$\\$0.000^{*}$} & \makecell{$+2.15$\\$8.0$\\$0.064$} & \makecell{$-9.70$\\$4.1$\\$0.001^{*}$} \\
LCG & \makecell{$+0.17$\\$6.0$\\$0.868$} & \makecell{$-4.16$\\$6.5$\\$0.005^{*}$} & \makecell{$-6.04$\\$6.5$\\$0.001^{*}$} & \makecell{$-11.92$\\$5.7$\\$0.000^{*}$} & \makecell{$-16.83$\\$7.5$\\$0.000^{*}$} & \makecell{$+0.35$\\$7.1$\\$0.734$} & \makecell{$-9.97$\\$4.1$\\$0.001^{*}$} \\
AugGPT & \makecell{$+0.84$\\$7.7$\\$0.425$} & \makecell{$-1.27$\\$8.0$\\$0.241$} & \makecell{$-6.56$\\$7.6$\\$0.000^{*}$} & \makecell{$-6.73$\\$7.2$\\$0.000^{*}$} & \makecell{$-12.78$\\$7.5$\\$0.000^{*}$} & \makecell{$+2.28$\\$6.9$\\$0.058$} & \makecell{$-6.15$\\$4.1$\\$0.003^{*}$} \\
CoTAM & \makecell{$-0.93$\\$5.7$\\$0.389$} & \makecell{$+0.01$\\$7.9$\\$0.995$} & \makecell{$-8.65$\\$6.8$\\$0.000^{*}$} & \makecell{$-4.10$\\$6.9$\\$0.005^{*}$} & \makecell{$-5.89$\\$7.1$\\$0.001^{*}$} & \makecell{$+3.53$\\$7.9$\\$0.008^{*}$} & \makecell{$-9.28$\\$4.1$\\$0.001^{*}$} \\
LLM2LLM & \makecell{$-2.09$\\$8.0$\\$0.071$} & \makecell{$+0.48$\\$6.2$\\$0.650$} & \makecell{$-7.52$\\$7.9$\\$0.000^{*}$} & \makecell{$-10.45$\\$5.1$\\$0.000^{*}$} & \makecell{$-7.07$\\$5.1$\\$0.001^{*}$} & \makecell{$+1.15$\\$6.8$\\$0.290$} & \makecell{$-21.04$\\$4.2$\\$0.000^{*}$} \\
CIEGAD & \makecell{$-0.87$\\$6.8$\\$0.412$} & \makecell{$-2.25$\\$4.9$\\$0.075$} & \makecell{$-6.25$\\$7.6$\\$0.000^{*}$} & \makecell{$-0.84$\\$6.8$\\$0.431$} & \makecell{$-9.19$\\$5.7$\\$0.000^{*}$} & \makecell{$-1.01$\\$5.6$\\$0.353$} & \makecell{$-4.70$\\$4.0$\\$0.009^{*}$} \\
VoidGen & \makecell{$-0.47$\\$5.7$\\$0.653$} & \makecell{$-2.37$\\$6.7$\\$0.051$} & \makecell{$-4.02$\\$7.8$\\$0.004^{*}$} & \makecell{$+2.12$\\$7.9$\\$0.068$} & \makecell{$-4.21$\\$5.3$\\$0.007^{*}$} & \makecell{$-0.31$\\$6.8$\\$0.764$} & N/A \\
\bottomrule
\end{tabular}
\end{table}
\section{Prompts}\label{app:prompts}
The exact prompt templates used for LLM-based augmentation methods are listed verbatim below.
Each method uses a fixed system message and a user message that is populated with dataset/class examples at generation time.
All templates are released with the source code; the CIEGAD-Judge template is summarized here for space; the full version is available in the released repository.

\paragraph{LLM-Paraphrase}
\begin{quote}\small\itshape
\textbf{System.} You are a helpful assistant that paraphrases text.
Keep the original meaning and emotional tone exactly. Output only
the paraphrased text, nothing else.\\
\textbf{User.} Paraphrase the following sentence while preserving
its meaning and emotion:\\
\{\textit{text}\}\\
Paraphrase:
\end{quote}

\paragraph{AugGPT (few-shot, ICL)}
\begin{quote}\small\itshape
\textbf{System.} You are a data augmentation assistant. Given a
few example sentences from a specific category, generate one new
sentence that belongs to the same category. The new sentence should
be diverse and different from the examples. Output only the new
sentence, nothing else.\\
\textbf{User.} Category: \{\textit{label\_name}\}\\
Here are \{\textit{n\_shots}\} example sentences from this category:\\
\{\textit{examples}\}\\
Generate one new sentence for the `\{\textit{label\_name}\}'
category (must be different from the examples above):
\end{quote}

\paragraph{Label-Conditioned Generation (LCG)}
\begin{quote}\small\itshape
\textbf{System.} You are a creative text generator. Given a target
emotion label and examples, generate a new, diverse text that
expresses that emotion naturally. Output only the generated text,
nothing else.\\
\textbf{User.} Generate a new text that expresses the emotion
`\{\textit{label\_name}\}'. Here are some examples of
`\{\textit{label\_name}\}' texts:\\
Example 1: \{\textit{ex\_1}\}\\
Example 2: \{\textit{ex\_2}\}\\
Example 3: \{\textit{ex\_3}\}\\
New `\{\textit{label\_name}\}' text (be diverse and creative):
\end{quote}

\paragraph{CIEGAD Generation (Interpolation and Extrapolation)}
For each cluster, CIEGAD constructs a profile card and performs two generation calls (one interpolation, one extrapolation).
The system prompt for the profile card is omitted for space; the generation prompt is reproduced in abbreviated form below.

\begin{quote}\small\itshape
\textbf{System.} You are a data generator that expands an English
emotion corpus.\\
\textbf{User.} [Cluster ID: $c_k$] [Emotion label:
\{\textit{label\_name}\}]\\
Cluster profile (summary): \{\textit{paragraph}\}\\
Cluster profile (JSON): \{\textit{json}\}\\
Inner examples (10): \{\textit{inner}\}\\
Outer examples (10): \{\textit{outer}\}\\
Task: \emph{Interpolation}---generate exactly 10 new English
sentences that remain faithful to the inner examples' emotional
tone, topical scope, and style, and explicitly avoid moving toward
the outer examples' direction. \emph{(Or, in extrapolation mode:
push beyond the outer examples' direction while not reverting
toward the inner centre.)}\\
Constraints: 10 items at once, distinct contexts, single sentences,
JSON output \texttt{[\{text, reason\}, $\dots$]}.
\end{quote}

\paragraph{CIEGAD Judge (LLM-as-a-Judge)}
The judge prompt scores each generated item on a five-point Likert scale across five aspects: emotion consistency, style alignment, lexical/topical coherence, contextual diversity, and reason validity.
Items falling below the per-aspect threshold are discarded.

\section*{Declaration of competing interest}
The author declares that he has no known competing financial interests or personal relationships that could have appeared to influence the work reported in this paper.

\section*{Acknowledgment}
This work was supported by JST SPRING, Grant Number JPMJSP2150.

\section*{Data availability}
All experimental code used in this study are publicly available at~\cite{void_aug_repo}. The seven public datasets used in this study are accessible via the HuggingFace Datasets hub under their original licenses.

\section*{Declaration of Generative AI Use}
During the preparation of this work, the author used large language models solely as a language-polishing aid for parts of the manuscript and for cross-checking English expressions. The author thereafter reviewed and edited the content as needed and takes full responsibility for the content of the publication. 

\bibliographystyle{elsarticle-num}
\bibliography{references}

\end{document}